%% file: main.tex
\documentclass[10pt]{article} 
\usepackage[preprint]{tmlr}

\input{math_commands.tex}

\usepackage{hyperref}
\usepackage{url}
\usepackage[utf8]{inputenc} 
\usepackage[T1]{fontenc}
\usepackage{booktabs}       
\usepackage{amsfonts}       
\usepackage{nicefrac}       
\usepackage{microtype}      
\usepackage{xcolor}
\usepackage{multirow}
\usepackage{graphicx}
\usepackage{subcaption}
\usepackage{amsmath}
\usepackage{enumitem}
\usepackage{algpseudocode}
\usepackage{algorithm}
\DeclareUnicodeCharacter{2212}{\textminus}
\usepackage{wrapfig}
\usepackage{bm}

\title{Continual Learning on a Data Diet}

\author{\name Elif Ceren Gok Yildirim \email e.c.gok@tue.nl \\
      \vspace{-6pt}
      \addr Eindhoven University of Technology
      \AND
      \name Murat Onur Yildirim \email m.o.yildirim@tue.nl \\
      \vspace{-6pt}
      \addr Eindhoven University of Technology
      \AND
      \name Joaquin Vanschoren \email j.vanschoren@tue.nl\\
      \vspace{-6pt}
      \addr Eindhoven University of Technology}

\def\month{MM}  
\def\year{YYYY} 
\def\openreview{\url{https://openreview.net/forum?id=XXXX}} 

\begin{document}
\maketitle

\vspace{-3pt}
\begin{abstract}
\vspace{-8pt}
Continual Learning (CL) methods usually learn from all the available data. However, this is not the case in human cognition which efficiently focuses on key experiences while disregarding the redundant information. Similarly, not all data points in a dataset have equal potential; some can be more informative than others. This disparity may significantly impact the performance, as both the quality and quantity of samples directly influence the model’s generalizability and efficiency. Drawing inspiration from this, we explore the potential of learning from important samples and present an empirical study for evaluating coreset selection techniques in the context of CL to stimulate research in this unexplored area. We train different continual learners on increasing amounts of selected samples and investigate the learning-forgetting dynamics by shedding light on the underlying mechanisms driving their improved stability-plasticity balance. We present several significant observations: learning from selectively chosen samples (i) enhances incremental accuracy, (ii) improves knowledge retention of previous tasks, and (iii) refines learned representations. This analysis contributes to a deeper understanding of selective learning strategies in CL scenarios. The code is available at \url{https://github.com/ElifCerenGokYildirim/DataDietCL}.
\end{abstract}

\input{1-intro}
\input{2-related_work}

\input{3-method}

\input{4-experimental_setting}

\input{5-Results}
\input{6-conclusion}

\subsubsection*{Broader Impact Statement}
This paper aims to advance the field of Machine Learning, especially on the subject of Class-Incremental Learning. Besides the advancements in the field, it shows training with smaller but more representative samples improves performance, thereby reducing memory and computation concerns.

\subsubsection*{Acknowledgement}
This work is supported by; TAILOR, a project funded by the EU Horizon 2020 research and innovation programme under GA No. 952215, Dutch national e-infrastructure with the support of SURF Cooperative using grant no. EINF-4569, and Turkish MoNE scholarship.


\bibliography{biblo}
\bibliographystyle{tmlr}

\newpage
\appendix
\section{Appendix}
\input{7-appendix}

\end{document}

%% file: math_commands.tex
\usepackage{amsmath,amsfonts,bm}

\def\eqref#1{equation~\ref{#1}}

\def\1{\bm{1}}

\DeclareMathAlphabet{\mathsfit}{\encodingdefault}{\sfdefault}{m}{sl}
\SetMathAlphabet{\mathsfit}{bold}{\encodingdefault}{\sfdefault}{bx}{n}



%% file: 1-intro.tex
\vspace{-10pt}
\section{Introduction}
\label{Intro}
\vspace{-5pt}
Humans exhibit a remarkable capacity to learn a multitude of tasks by progressively accumulating knowledge and skills over time.
Continual Learning (CL) mimics this ability and aims to sequentially learn from a stream of data while retaining previously acquired knowledge. Class-Incremental Learning (CIL) is the most challenging scenario where the learner is required to predict outcomes for all encountered classes without being given task identifiers~\citep{CLscnearios}.
However, catastrophic forgetting~\citep{catastrophicforgetting} remains a challenge in this dynamic setting wherein the class-incremental learners tend to lose acquired knowledge from previous tasks, upon learning new ones. Recent research has brought solutions through various techniques, including  regularization methods~\citep{ewc, lwf, imm}, replay strategies~\citep{agem, gem, gss, coresetsviabilevel}, architecture expansion~\citep{der, foster, memo, pnn, scalable} and prompt learning~\citep{l2p, dualprompt, coda-prompt} approaches. However, these approaches aim to learn from all the available data during training to maximize model performance and assume that all samples are equally important. This standardized practice may not fully reflect the efficiency and adaptability observed in human learning since, as humans, we intuitively filter and prioritize information, focusing on key experiences (e.g. clear and novel examples) that enrich our understanding while disregarding redundant details~\citep{selective, cognitive, brain_attention}.

\begin{figure*}[t]
  \centering
  \includegraphics[width=\textwidth]{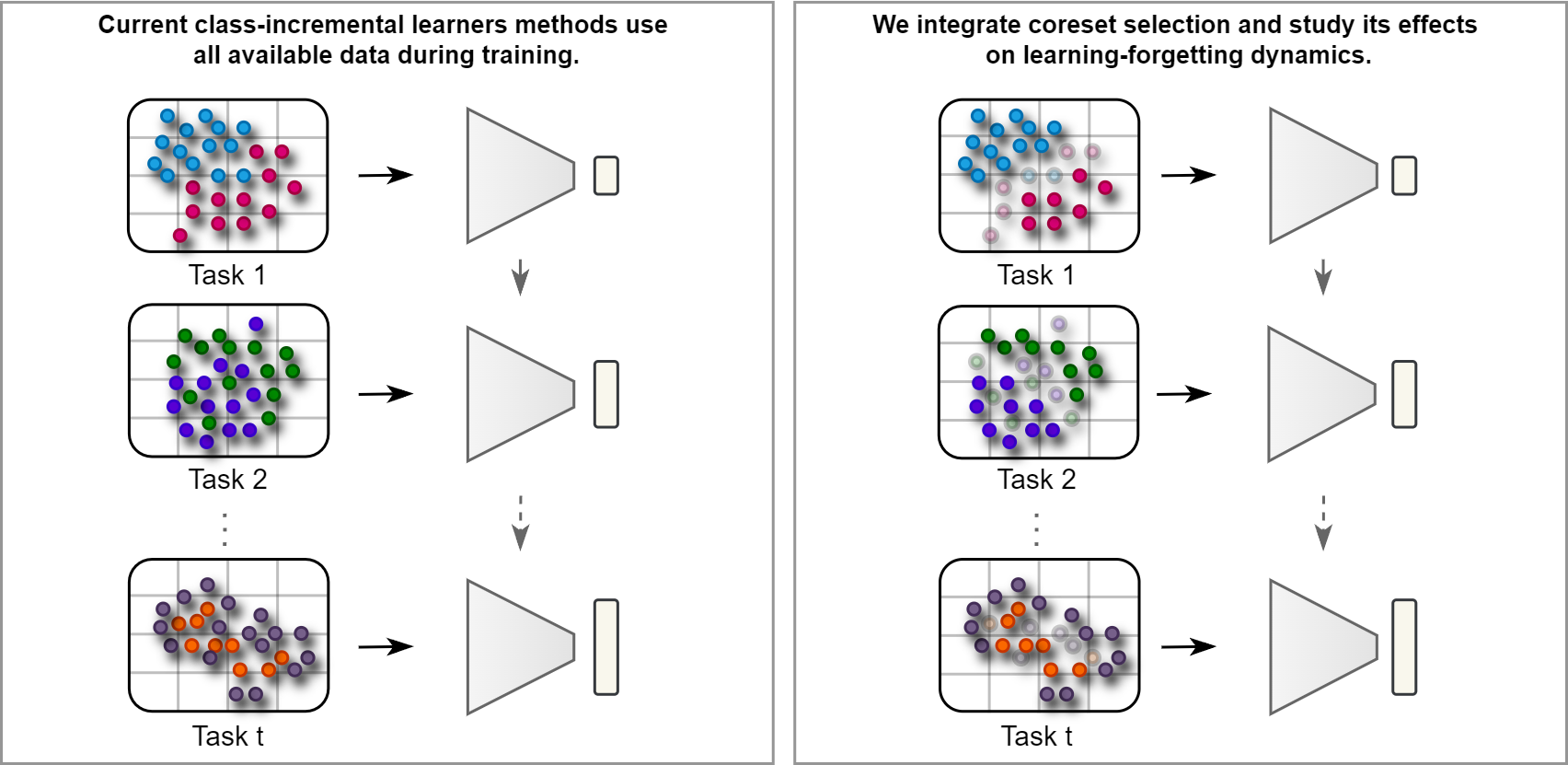}
    \caption{Illustration of our evaluation protocol: Existing class-incremental learning methods (\textbf{left}) typically utilize all available samples indiscriminately during training. In this study (\textbf{right}), we subject class-incremental learners to a \textit{data diet} and analyze how the selection of the most important samples with different coreset selection methods affects the incremental performance.}
  \label{fig:fig1}
  \vskip -0.2cm
\end{figure*}

We draw inspiration from this human cognitive ability and introduce an empirical study to evaluate the learning-forgetting dynamics of different CIL models when trained with important samples selected by a wide range of sample selection approaches (as illustrated in Figure \ref{fig:fig1}).
Through a detailed analysis, we provide insight into the underlying reasons of stability-plasticity balance. We believe that this comprehensive study and investigation contributes to a deeper understanding of the potential benefits of sample selective learning strategies in CIL scenarios and stimulates systematic research that aims not only to refine model structures but also to prioritize the inherent quality of the data in the learning process.  

\newpage
Our contributions can be summarized as:
\begin{enumerate}[label=\Roman*.]
    \item This paper presents the first explicit empirical analysis of different coreset selection methods in combination with various continual learners in the class-incremental learning setting.
    \item We find that learning from selectively chosen samples with different coreset selection methods significantly elevates incremental performance.
    \item We demonstrate that the increase in performance among class-incremental learners trained with selected samples arises from enhanced retention of previously acquired concepts due to improved representation and perception power of the models.
    \item We show that continual learning can benefit from a data-centric approach, despite the fact that most existing research has predominantly focused on model-centric enhancements.
\end{enumerate}

%% file: 2-related_work.tex
\section{Background}
\label{Literature}
\subsection{Class-Incremental Learning.}

Class-incremental learning can be broadly categorized into three main approaches~\citep{CLscnearios}; regularization, replay, architecture-based and prompt-based. Regularization-based methods regularize the abrupt changes in the learned parameters to prevent catastrophic forgetting~\citep{ewc, lwf, imm}. Replay-based methods either retain selected exemplars from prior tasks or generate a subset of data points from previous tasks to alleviate forgetting~\citep{agem, gem, gss, coresetsviabilevel}. Architecture-based methods prevent forgetting by increasing model size and allocating distinct sets of parameters to individual tasks, ensuring there is no overlap between them~\citep{der,foster,memo,pnn,scalable}.Recently, with the growing popularity of large pretrained models with Vision Transformers (ViT), prompt-based methods also received growing popularity~\citep{l2p,dualprompt,coda-prompt}.

\subsection*{Summary of CIL Methods Selected for Analysis}
\label{section 2.1}
We use 7 well-established CIL models that encompass various approaches including architecture-based, replay-based, regularization-based, and prompt-based. We deliberately chose these methods to provide a comprehensive analysis since they all represent different learning strategies.

\paragraph{DER-Architecture.}
Dynamically expandable representation \citep{der} creates a new backbone for each task and then aggregates the features of all backbones on a single classifier. Each new or expanded backbone uses an additional auxiliary loss to differentiate better between old and new classes. When facing new tasks, it freezes the old backbone to maintain former knowledge.

\paragraph{FOSTER-Architecture.} 
Feature boosting and compression for class-incremental learning \citep{foster} frames the learning process as a feature-boosting problem and aims to enhance the learning of new features. Then, it expands the continual learner on a single classifier by integrating the boosted features with a compression step to ensure that only relevant features are retained.

\paragraph{MEMO-Architecture.}
Memory efficient expandable model \citep{memo} expands the network in a more efficient way. It assumes that the initial blocks of the backbone capture the general patterns for any task and only expands the model in the last or specialized blocks that are designed to be task-specific.

\paragraph{iCaRL-Replay.}
Incremental Classifier and Representation Learning \citep{icarl} is a replay-based method that stores samples from each learned task. Upon the arrival of a new task, it uses stored exemplars together with the new one to capture the distribution at once.
Therefore, it refines the features after each task with additional distillation loss to overcome abrupt shifts in the feature space.

\paragraph{ER-Replay.}
Experience Replay \citep{er} is a simple yet strong method that employs reservoir sampling to store samples from each task and randomly retrieves stored samples with the new task to capture the distribution all at once. 

\paragraph{LwF-Regularization.}
Learning without Forgetting \citep{lwf} is solely a regularization-based method without relying on any replay buffer. It utilizes a distillation loss to prevent sudden changes in the feature space while learning new tasks.

\paragraph{CODA-Prompt.}
CODA-Prompt \citep{coda-prompt} as the name suggested is a prompt based method that leverages pretrained Vision Transformers (ViT) without relying on data rehearsal. It introduces a set of prompt components that are dynamically assembled based on input-conditioned weights, generating task-specific prompts for the transformer’s attention layers. These generated prompts selectively guide the model’s attention to relevant features for each task, to enable better stability-plasticity tradeoff. 

\subsection{Coreset Selection.}
Coreset selection approximates the distribution of the whole dataset with a small subset and has been extensively examined in data-efficient supervised batch learning~\citep{forgetting, deepcore, efficientdataselection, datadiet, herding, uncertainty, graphcut, dataefficinetcoreset} and active learning~\citep{coresetonactivelearning, coresetonactivelearning2}. Coreset selection also holds promise in continual learning to construct a memory buffer from important samples~\citep{gss, coresetsviabilevel}. Recently, an inspiring study~\citep{coreset} improved the performance in online CL setup by introducing a coreset selection method to select the most diverse samples while approximating the mean of a given batch.

However, besides this one method~\citep{coreset}, the interplay between coreset selection methods and continual learning models remains unexplored. This warrants deeper investigation into their interaction as well as the underlying mechanisms related to the improved performance. Exploring this interaction, by focusing on the quality of the data itself, could provide novel insight to create more efficient and advanced continual learners. 

\subsubsection*{Overview of Coreset Algorithms Selected for Analysis}
\label{section 2.2}
We employ 4 distinct coreset selection methods as well as a baseline using random selection. Once again, we carefully chose these distinct methods to offer comprehensive empirical analysis. It is important to note that these coreset selection methods require a brief initial training or warm-up phase to make informed and meaningful decisions when selecting coreset samples.

\paragraph{Random.}
This selection strategy involves randomly selecting a subset of data points from the entire dataset without any specific criteria or consideration of their importance or informativeness.

\paragraph{Herding.}
Herding \citep{herding} chooses data points by evaluating the distance between the center of the original dataset and the center of the coreset within the feature space. This algorithm progressively and greedily includes one sample at a time into the coreset, aiming to minimize the distance between centers.

\paragraph{Uncertainty.}
Samples with lower confidence levels might have a stronger influence than those with higher confidence levels, thus having these samples in the coreset can be useful. Least confidence, entropy, and margin are the common metrics used to quantify sample uncertainty \citep{uncertainty}. In this study, entropy is used as a selection metric.

\paragraph{Forgetting.}
Forgetting selects instances which were correctly classified in one epoch and then subsequently misclassified in the following epoch during training \citep{forgetting}. This method provides valuable insight into the intrinsic characteristics of the training data and removes challenging or forgettable instances.

\paragraph{GraphCut.}
GraphCut partitions the dataset into subsets based on dissimilarity or information content, and data points from these subsets are then selected to form the coreset \citep{graphcut}. This approach ensures that the coreset captures the diversity and essential information of the original dataset while reducing redundancy.

%% file: 3-method.tex
\section{Data Diet}
\label{Method}

We conduct a comprehensive evaluation of existing CL methods, assessing their performance when trained on a reduced set of informative samples, as opposed to the traditional approach of full dataset training. We refer to this as a `Data Diet'.
To clarify our approach, we first present the necessary preliminaries and problem formulation in Section \ref{section 3.1}. Following this, we define our objective and outline the proposed training strategy in Section \ref{section 3.2}.

\subsection{Preliminaries and Problem Formulation}
\label{section 3.1}
Formally, we define the CIL problem as a sequence of classification tasks $T_{1:t} = (T_1, T_2, ..., T_t)$. Each task $T_t$ is drawn from an unknown distribution and consists of input pairs ($x_{i,t}$, $y_{i,t}) \in X_t \times Y_t$ where $x_{i,t}$ represents the sample and $y_{i,t}$ indicates the corresponding label. Note that, these learning tasks are mutually exclusive, meaning that the label sets do not overlap, i.e., $Y_{t-1} \cap Y_t = \emptyset$. 

From the coreset selection perspective, the aim is to find
the most informative subset $S_t$ from a given task $T_t$ with a large number of input pairs ($x_{i,t}$, $y_{i,t}$). Therefore, model trained with subset $S_t \subset T_t$ with a condition of $|S_t| < |T_t|$ has a close generalization performance compared to a model trained with $T_t$. 

\subsection{Objective and Training Strategy}
\label{section 3.2}
We train the continual learning models within a specified time budget for each task, aiming to minimize their loss function. We structure the training process into two distinct phases: the warm-up phase and the learning phase. This is necessary because coreset selection methods require an initial warm-up phase to identify the most informative samples correctly. It is important to note that the duration of the warm-up phase is typically much shorter than that of the learning phase. Upon completion of the warm-up phase, the learning phase proceeds with the selected subset of samples.

\begin{algorithm}
\caption{CL on Data Diet}
\label{algo}
\begin{algorithmic}
\Require Model $f_\theta$, Tasks $T_{1:t}$ with training sets $T_t$, learning rate $\eta$, total epochs $e$, warm-up fraction $\alpha \in (0,1)$, coreset selector $\phi$, coreset fraction  $s \in (0,1)$
\For{task t = 1 to $\lfloor T \rfloor$}
\For{epoch = 1 to $\lfloor\alpha e\rfloor$} \Comment{Warm-up Phase}
    \For{each batch $b$ in $T_t$}
        \State Compute $\mathcal{L}_{CE}(f_\theta, b)$
        \State Update $f_\theta \gets \theta - \eta \nabla_\theta \mathcal{L}_{CE}$
    \EndFor
\EndFor
\State Use $\phi$ to select $S_t \subset T_t$ with a fraction of $s$ 
\For{epoch = 1 to $\lfloor(1-\alpha)e\rfloor$} \Comment{Learning Phase}
    \For{each batch $b$ in $S_t$}
        \State Compute $\mathcal{L}_{\text{CL}}(f_\theta, b)$
       \State Update $f_\theta \gets \theta - \eta \nabla_\theta \mathcal{L}_{CL}$
    \EndFor
\EndFor
\EndFor
\end{algorithmic}
\end{algorithm}

Let $f_{\theta}(\cdot)$ denote the continual learning model with parameters $\theta$. Then, the training process can then be expressed as follows:

\begin{align}
\label{equ:training}
f_{\theta^*} = \arg\min_\theta \mathcal{L}_{CL}(f_\theta, S_t, (1-\alpha)e) \circ \arg\min_\theta \mathcal{L}_{CE}(f_\theta, T_t, \alpha e)
\end{align}

Here, $(f_\theta, T_t, \alpha e)$ represents training the model $f_\theta$ on the full task $T_t$ with a defined time budget of $\alpha e$ where hyperparameter $\alpha \in (0,1)$ and determines the fraction of the total training budget allocated to the warm-up phase, and $e$ is the total number of epochs available for training.
Similarly, $(f_\theta, S_t, (1-\alpha)e)$ represents the training of the model $f_\theta$, for the remaining time budget $(1- \alpha)e$, on the coreset $S_t$ which is selected from $T_t$ with a fraction of $s\in (0,1)$ by a coreset selector $\phi$, so that $|S_t| = s \cdot |T_t|$. Note that, $\mathcal{L}_{CE}$ represents Cross-Entropy loss and $\mathcal{L}_{CL}$ represents the loss defined by continual learning methods given in section \ref{section 2.1}. We give the detailed flow of the CL on Data Diet in Algorithm \ref{algo}.

%% file: 4-experimental_setting.tex
\vspace{-3pt}
\section{Experimental Setting}
\label{Experimental Setting}

\paragraph{Datasets.}
We use well-established continual learning datasets, specifically \textbf{Split-CIFAR10} and \textbf{Split-CIFAR100}~\citep{cifar}, \textbf{ImageNet-100}~\citep{imagenet} in our experiments to evaluate and posit our findings.
\textbf{Split-CIFAR10} has 5 disjoint tasks and each task has 2 disjoint classes with 10000 samples for training and 2000 samples for testing. \textbf{Split-CIFAR100} has 10 disjoint tasks and each task has 10 disjoint classes with 5000 samples for training and 1000 samples for testing. In addition, we employ \textbf{Split-ImageNet100}, a subset of the large-scale ImageNet dataset, with images at a higher resolution of 224x224 pixels. Similar to Split-CIFAR100, Split-ImageNet100 is divided into 10 tasks, each consisting of 10 disjoint classes. The increased number of classes, fewer images per class combined with longer learning sessions, and higher resolution bring further challenges and offer a more complex scenario. 

\paragraph{Backbones.}
To offer a more comprehensive evaluation, we test both from scratch and pretrained models across two architectures: ResNet18 \citep{resnet18} and Vision Transformer (ViT) \citep{vit}. In \textbf{ResNet18} trained from scratch, we observe how well it can learn task-specific features directly from the dataset. In contrast, the pretrained models \textbf{Pretrained-ResNet18} and \textbf{Pretrained-ViT} are initialized with ImageNet weights, giving them prior knowledge of visual patterns and structures, which helps them start with a robust foundation for CL. Results for the pretrained models are provided in the Appendix.

\paragraph{Metrics.}
We utilize standard CL evaluation metrics, including average accuracy (ACC) and backward transfer (BWT). The average accuracy metric measures the final accuracy averaged over all tasks and can be formulated as $ACC = \frac{1}{T} \sum_{i=1}^{T} A_{T,i}$ where \(A_{T,i}\) represents the testing accuracy of task \(T\) after learning task \(i\). Backward transfer metric quantifies the forget by measuring the accuracy change of each task after learning new ones and can be formally represented as $BWT = \frac{1}{T-1} \sum_{i=1}^{T-1} (A_{T,i} - A_{i,i})$.

\paragraph{Implementation Details.}
We use Deepcore \citep{deepcore} for coreset selection methods and PYCIL \citep{pycil} for the class-incremental learners. We set the total training budget $e$ to 100 epochs where warmup fraction $\alpha$ is set to $0.1$ and the remaining is allocated for the learning phase. We set coreset fraction $s$ to $10\%, 20\%, 50\%, 80\%$ and $90\%$ for each task. We use a Stochastic Gradient Descent (SGD) optimizer with a scheduled learning rate of 0.1 and momentum of 0.9. We set a weight decay of $5 \times 10^{-4}$ for the initial task and $2 \times 10^{-4}$ for subsequent tasks. We set the batch size to 128. We employ a fixed memory size: 50 per class for Split-CIFAR10 and 20 per class for Split-CIFAR100 and Split-ImageNet100. For ViT training, we only modify the learning rate to 0.001, reduce the batch size to 32, and train for 20 epochs. We run experiments on Nvidia A100 GPU with different seeds and report the average across three runs.

%% file: 5-Results.tex
\section{Results and Analysis}
\label{Results}
In Section~\ref{section 5.1}, we conduct a thorough analysis across diverse CIL methods and different coreset selection algorithms with varying coreset sizes.
In Section~\ref{section 5.2}, we investigate why coreset selection improves incremental accuracy, offering insight into the stability-plasticity dynamics of each class-incremental learner. 
In Section~\ref{section 5.3}, we seek to understand how these dynamics are reflected in the learning perception of the model.

\begin{table}[h]
\caption{Accuracy [\%] of CIL models across various coreset fractions and selections on \textbf{Split-CIFAR10}. Learning from coreset samples enhances the performance, except FOSTER and LwF. The best results are highlighted in bold if coreset selection outperforms training with all samples.}
\label{tab:cifar10}
\resizebox{\columnwidth}{!}{%
\begin{tabular}{llllllll}
\hline
 &
  Fraction &
  \multicolumn{1}{c}{10\%} &
  \multicolumn{1}{c}{20\%} &
  \multicolumn{1}{c}{50\%} &
  \multicolumn{1}{c}{80\%} &
  \multicolumn{1}{c}{90\%} &
  \multicolumn{1}{c}{100\%} \\ \hline
\multicolumn{1}{l}{\multirow{5}{*}{DER~\citep{der}}} &
  \multicolumn{1}{l}{Random} &
  \multicolumn{1}{c}{51.79 ± 4.6} &
  \multicolumn{1}{c}{54.28 ± 3.8} &
  \multicolumn{1}{c}{55.68 ± 0.3} &
  \multicolumn{1}{c}{57.27 ± 2.9} &
  \multicolumn{1}{c}{55.61 ± 2.5} &
  \multicolumn{1}{c}{56.91 ± 1.3} \\
\multicolumn{1}{c}{} &
  \multicolumn{1}{l}{Herding} &
  41.65 ± 2.2 &
  52.35 ± 2.5 &
  59.79 ± 1.8 &
  63.96 ± 1.1 &
  62.93 ± 1.2 &
  56.91 ± 1.3 \\
\multicolumn{1}{c}{} &
  \multicolumn{1}{l}{Uncertainty} &
  56.02 ± 1.7 &
  59.48 ± 1.7 &
  57.97 ± 0.8 &
  62.01 ± 3.1 &
  59.36 ± 1.5 &
  56.91 ± 1.3 \\
\multicolumn{1}{c}{} &
  \multicolumn{1}{l}{Forgetting} &
  55.68 ± 2.1 &
  60.97 ± 1.0 &
  60.82 ± 0.3 &
  63.46 ± 3.9 &
  61.36 ± 0.4 &
  56.91 ± 1.3 \\
\multicolumn{1}{c}{} &
  \multicolumn{1}{l}{GraphCut} &
  62.06 ± 1.9 &
  \textbf{64.74 ± 0.5} &
  63.03 ± 2.0 &
  61.17 ± 1.9 &
  62.95 ± 1.5 &
  56.91 ± 1.3 \\ \hline
\multicolumn{1}{l}{\multirow{5}{*}{FOSTER~\citep{foster}}} &
  \multicolumn{1}{l}{Random} &
  52.44 ± 5.4 &
  52.34 ± 4.3 &
  53.22 ± 2.8 &
  53.93 ± 4.2 &
  53.93 ± 3.0 &
  54.79 ± 2.9 \\
\multicolumn{1}{l}{} &
  \multicolumn{1}{l}{Herding} &
  32.00 ± 2.2 &
  39.91 ± 8.3 &
  46.91 ± 3.3 &
  52.82 ± 2.6 &
  51.34 ± 1.2 &
  54.79 ± 2.9 \\
\multicolumn{1}{l}{} &
  \multicolumn{1}{l}{Uncertainty} &
  45.42 ± 3.6 &
  49.18 ± 4.6 &
  48.94 ± 3.2 &
  50.95 ± 2.6 &
  49.25 ± 2.2 &
  54.79 ± 2.9 \\
\multicolumn{1}{l}{} &
  \multicolumn{1}{l}{Forgetting} &
  45.44 ± 3.2 &
  51.59 ± 4.0 &
  49.37 ± 0.2 &
  48.19 ± 2.6 &
  49.10 ± 1.5 &
  54.79 ± 2.9 \\
\multicolumn{1}{l}{} &
  \multicolumn{1}{l}{GraphCut} &
  50.85 ± 3.1 &
  52.54 ± 3.7 &
  49.94 ± 0.3 &
  49.43 ± 0.9 &
  49.28 ± 1.0 &
  54.79 ± 2.9 \\ \hline
\multicolumn{1}{l}{\multirow{5}{*}{MEMO~\citep{memo}}} &
  \multicolumn{1}{l}{Random} &
  44.36 ± 4.2 &
  45.41 ± 5.5 &
  47.45 ± 6.4 &
  48.93 ± 7.1 &
  49.58 ± 7.2 &
  49.22 ± 5.5 \\
\multicolumn{1}{l}{} &
  \multicolumn{1}{l}{Herding} &
  39.32 ± 0.2 &
  45.04 ± 0.4 &
  47.90 ± 3.1 &
  49.98 ± 6.1 &
  49.34 ± 6.3 &
  49.22 ± 5.5 \\
\multicolumn{1}{l}{} &
  \multicolumn{1}{l}{Uncertainty} &
  38.27 ± 6.9 &
  41.10 ± 5.0 &
  44.99 ± 6.4 &
  47.75 ± 6.0 &
  47.90 ± 5.4 &
  49.22 ± 5.5 \\
\multicolumn{1}{l}{} &
  \multicolumn{1}{l}{Forgetting} &
  35.04 ± 4.1 &
  45.23 ± 5.4 &
  47.74 ± 5.3 &
  48.66 ± 5.5 &
  47.78 ± 5.9 &
  49.22 ± 5.5 \\
\multicolumn{1}{l}{} &
  \multicolumn{1}{l}{GraphCut} &
  51.37 ± 3.6 &
  \textbf{52.54 ± 2.3} &
  49.67 ± 4.0 &
  49.97 ± 6.0 &
  48.35 ± 5.7 &
  49.22 ± 5.5 \\ \hline
\multicolumn{1}{l}{\multirow{5}{*}{iCaRL~\citep{icarl}}} &
  \multicolumn{1}{l}{Random} &
  47.70 ± 4.3 &
  55.41 ± 5.4 &
  54.56 ± 5.8 &
  57.75 ± 7.5 &
  57.29 ± 6.3 &
  59.54 ± 8.0 \\
\multicolumn{1}{l}{} &
  \multicolumn{1}{l}{Herding} &
  40.32 ± 5.0 &
  42.99 ± 3.3 &
  54.02 ± 4.5 &
  58.60 ± 6.7 &
  59.11 ± 6.3 &
  59.54 ± 8.0 \\
\multicolumn{1}{l}{} &
  \multicolumn{1}{l}{Uncertainty} &
  50.77 ± 1.5 &
  54.41 ± 6.2 &
  56.78 ± 6.3 &
  57.38 ± 6.6 &
  57.82 ± 7.1 &
  59.54 ± 8.0 \\
\multicolumn{1}{l}{} &
  \multicolumn{1}{l}{Forgetting} &
  53.79 ± 4.9 &
  57.86 ± 5.9 &
  58.30 ± 5.9 &
  58.90 ± 6.3 &
  56.90 ± 7.7 &
  59.54 ± 8.0 \\
\multicolumn{1}{l}{} &
  \multicolumn{1}{l}{GraphCut} &
  \textbf{61.70 ± 2.7} &
  61.07 ± 4.2 &
  60.88 ± 5.6 &
  58.80 ± 7.0 &
  57.68 ± 7.1 &
  59.54 ± 8.0 \\ \hline
\multicolumn{1}{l}{\multirow{5}{*}{ER~\citep{er}}} &
  \multicolumn{1}{l}{Random} &
  51.02 ± 2.7 &
  56.32 ± 6.2 &
  57.79 ± 4.6 &
  57.20 ± 6.0 &
  57.77 ± 6.9 &
  58.51 ± 6.4 \\
\multicolumn{1}{l}{} &
  \multicolumn{1}{l}{Herding} &
  41.06 ± 7.5 &
  47.97 ± 4.0 &
  55.87 ± 4.9 &
  58.93 ± 4.6 &
  58.85 ± 4.9 &
  58.51 ± 6.4 \\
\multicolumn{1}{l}{} &
  \multicolumn{1}{l}{Uncertainty} &
  52.70 ± 2.4 &
  52.99 ± 1.1 &
  56.35 ± 6.3 &
  57.48 ± 6.4 &
  58.09 ± 5.4 &
  58.51 ± 6.4 \\
\multicolumn{1}{l}{} &
  \multicolumn{1}{l}{Forgetting} &
  52.44 ± 3.4 &
  55.05 ± 5.8 &
  57.43 ± 5.7 &
  57.00 ± 5.5 &
  56.73 ± 6.2 &
  58.51 ± 6.4 \\
\multicolumn{1}{l}{} &
  \multicolumn{1}{l}{GraphCut} &
  \textbf{63.03 ± 3.1} &
  60.53 ± 2.6 &
  60.34 ± 4.4 &
  58.69 ± 5.6 &
  57.61 ± 5.8 &
  58.51 ± 6.4 \\ \hline
\multicolumn{1}{l}{\multirow{5}{*}{LwF~\citep{lwf}}} &
  \multicolumn{1}{l}{Random} &
  31.60 ± 0.8 &
  41.46 ± 1.9 &
  45.64 ± 1.5 &
  51.21 ± 4.7 &
  \textbf{51.83 ± 2.1} &
  51.15 ± 4.3 \\
\multicolumn{1}{l}{} &
  \multicolumn{1}{l}{Herding} &
  15.27 ± 3.8 &
  23.75 ± 3.0 &
  20.72 ± 0.7 &
  27.74 ± 5.2 &
  30.86 ± 4.1 &
  51.15 ± 4.3 \\
\multicolumn{1}{l}{} &
  \multicolumn{1}{l}{Uncertainty} &
  26.89 ± 5.0 &
  24.21 ± 3.3 &
  28.95 ± 5.1 &
  29.58 ± 5.8 &
  30.54 ± 4.2 &
  51.15 ± 4.3 \\
\multicolumn{1}{l}{} &
  \multicolumn{1}{l}{Forgetting} &
  27.10 ± 5.3 &
  25.49 ± 4.0 &
  27.66 ± 5.2 &
  30.24 ± 5.5 &
  30.57 ± 5.0 &
  51.15 ± 4.3 \\
\multicolumn{1}{l}{} &
  \multicolumn{1}{l}{GraphCut} &
  25.34 ± 3.1 &
  26.22 ± 3.5 &
  29.42 ± 5.2 &
  30.54 ± 4.2 &
  30.95 ± 5.4 &
  51.15 ± 4.3 \\ \hline
\end{tabular}%
}
\end{table}

\subsection{Data diet enhances incremental performance}
\label{section 5.1}

\paragraph{Large number of samples per task.}
Our analysis reveals a consistent trend of performance enhancement across various class-incremental learners when utilizing coreset selection strategies (see Table \ref{tab:cifar10}). 
We find that when the coreset size is large enough, all selection methods tend to exhibit comparable performance. Conversely, in scenarios where the coreset size is more restricted, a sophisticated method like GraphCut outperforms others.
Moreover, the size of the coreset also plays a role: smaller coresets tend to yield more significant improvements due to increased distinction between representations which we discuss more in detail in Section \ref{section 5.3}. This observation is particularly evident in the case of DER which demonstrates a remarkable enhancement of approximately 7\% in performance when trained only with 20\% of the samples from each task. Finally, we observe that the benefit of coreset selection on FOSTER and LwF appears less pronounced.

\paragraph{Small number of samples per task.}
When the number of samples per task is relatively limited, we still observe performance enhancements, although they are not as pronounced due to the increased challenge of selecting informative samples (see in Table \ref{tab:cifar100} and \ref{tab:imagenet_subset}). Consequently, in such situations, opting for a larger coreset is more beneficial since a smaller coreset size would result in an exceptionally small sample size per task, posing a challenge for class-incremental learners. For instance, in Table \ref{tab:cifar100}, iCaRL improves its performance by around 3\% when trained with 80\% of the samples from each task, compared to full sample training. However, its performance stars to degrade when coreset size is less than 50\%.

\paragraph{Experiments on pretrained backbone.}
We further complemented our study with pretrained ResNet18 and ViT backbones where the results align with the findings discussed herein. We observe that pretraining improves the performance regardless of coreset selection. However, coreset selection provides an additional performance boost. For more details, please refer to the Appendix \ref{appendix-pretrained}. 

\begin{table}[h]
\caption{Accuracy [\%] of CIL models across various coreset fractions and selections on \textbf{Split-CIFAR100}. Learning from coreset samples enhances the performance, except FOSTER and LwF. The best results are highlighted in bold if coreset outperforms training with all samples.}
\label{tab:cifar100}
\resizebox{\columnwidth}{!}{%
\begin{tabular}{llllllll}
\hline
 &
  Fraction &
  \multicolumn{1}{c}{10\%} &
  \multicolumn{1}{c}{20\%} &
  \multicolumn{1}{c}{50\%} &
  \multicolumn{1}{c}{80\%} &
  \multicolumn{1}{c}{90\%} &
  \multicolumn{1}{c}{100\%} \\ \hline
\multicolumn{1}{l}{\multirow{5}{*}{DER~\citep{der}}} &
  \multicolumn{1}{l}{Random} &
  26.23 ± 0.6 &
  36.35 ± 2.8 &
  47.32 ± 2.6 &
  53.11 ± 1.6 &
  54.07 ± 0.1 &
  53.81 ± 1.0 \\
\multicolumn{1}{c}{} &
  \multicolumn{1}{l}{Herding} &
  17.99 ± 7.5 &
  24.79 ± 6.0 &
  41.11 ± 2.7 &
  52.48 ± 0.4 &
  53.92 ± 0.8 &
  53.81 ± 1.0 \\
\multicolumn{1}{c}{} &
  \multicolumn{1}{l}{Uncertainty} &
  27.54 ± 4.6 &
  38.29 ± 3.0 &
  49.41 ± 1.2 &
  \textbf{55.71 ± 1.9} &
  54.55 ± 0.4 &
  53.81 ± 1.0 \\
\multicolumn{1}{c}{} &
  \multicolumn{1}{l}{Forgetting} &
  30.32 ± 4.9 &
  41.25 ± 1.8 &
  49.20 ± 2.2 &
  54.10 ± 0.3 &
  53.68 ± 0.1 &
  53.81 ± 1.0 \\
\multicolumn{1}{c}{} &
  \multicolumn{1}{l}{GraphCut} &
  29.61 ± 5.7 &
  39.71 ± 3.4 &
  50.35 ± 1.0 &
  53.08 ± 0.8 &
  54.89 ± 0.7 &
  53.81 ± 1.0 \\ \hline
\multicolumn{1}{l}{\multirow{5}{*}{FOSTER~\citep{foster}}} &
  \multicolumn{1}{l}{Random} &
  23.21 ± 0.0 &
  32.04 ± 1.3 &
  48.95 ± 0.8 &
  51.71 ± 1.9 &
  53.34 ± 0.8 &
  56.19 ± 2.3 \\
\multicolumn{1}{l}{} &
  \multicolumn{1}{l}{Herding} &
  10.84 ± 0.8 &
  18.38 ± 1.1 &
  35.15 ± 2.7 &
  51.51 ± 0.1 &
  53.72 ± 0.9 &
  56.19 ± 2.3 \\
\multicolumn{1}{l}{} &
  \multicolumn{1}{l}{Uncertainty} &
  16.97 ± 0.1 &
  27.37 ± 0.9 &
  44.29 ± 3.1 &
  55.24 ± 0.1 &
  55.10 ± 1.7 &
  56.19 ± 2.3 \\
\multicolumn{1}{l}{} &
  \multicolumn{1}{l}{Forgetting} &
  21.80 ± 0.4 &
  32.42 ± 0.8 &
  44.97 ± 2.9 &
  54.59 ± 0.4 &
  54.91 ± 1.0 &
  56.19 ± 2.3 \\
\multicolumn{1}{l}{} &
  \multicolumn{1}{l}{GraphCut} &
  22.16 ± 1.6 &
  30.40 ± 1.1 &
  45.91 ± 2.3 &
  53.35 ± 1.9 &
  55.24 ± 0.5 &
  56.19 ± 2.3 \\ \hline
\multicolumn{1}{l}{\multirow{5}{*}{MEMO~\citep{memo}}} &
  \multicolumn{1}{l}{Random} &
  20.79 ± 0.7 &
  26.74 ± 0.1 &
  29.62 ± 0.5 &
  34.58 ± 0.1 &
  34.58 ± 0.1 &
  34.23 ± 0.4 \\
\multicolumn{1}{l}{} &
  \multicolumn{1}{l}{Herding} &
  13.24 ± 2.0 &
  18.76 ± 1.5 &
  27.26 ± 1.8 &
  33.64 ± 0.3 &
  \textbf{34.94 ± 0.1} &
  34.23 ± 0.4 \\
\multicolumn{1}{l}{} &
  \multicolumn{1}{l}{Uncertainty} &
  16.07 ± 2.6 &
  23.23 ± 2.9 &
  30.14 ± 1.7 &
  33.41 ± 0.9 &
  34.10 ± 1.0 &
  34.23 ± 0.4 \\
\multicolumn{1}{l}{} &
  \multicolumn{1}{l}{Forgetting} &
  18.44 ± 1.9 &
  23.37 ± 2.0 &
  31.17 ± 0.3 &
  33.10 ± 0.4  &
  32.46 ± 2.2 &
  34.23 ± 0.4 \\
\multicolumn{1}{l}{} &
  \multicolumn{1}{l}{GraphCut} &
  23.21 ± 1.7 &
  27.79 ± 0.6 &
  32.49 ± 0.6 &
  33.61 ± 0.2 &
  34.22 ± 0.7 &
  34.23 ± 0.4 \\ \hline
\multicolumn{1}{l}{\multirow{5}{*}{iCaRL~\citep{icarl}}} &
  \multicolumn{1}{l}{Random} &
  25.48 ± 0.2 &
  29.87 ± 3.0 &
  35.37 ± 2.0 &
  37.02 ± 3.1 &
  37.11 ± 3.0 &
  37.45 ± 1.7 \\
\multicolumn{1}{l}{} &
  \multicolumn{1}{l}{Herding} &
  13.02 ± 1.2 &
  17.24 ± 1.5 &
  27.91 ± 1.3 &
  38.24 ± 1.3 &
  37.55 ± 0.8 &
  37.45 ± 1.7 \\
\multicolumn{1}{l}{} &
  \multicolumn{1}{l}{Uncertainty} &
  22.47 ± 1.9 &
  28.05 ± 1.3 &
  35.18 ± 3.3 &
  \textbf{40.25 ± 0.7} &
  39.26 ± 2.5 &
  37.45 ± 1.7 \\
\multicolumn{1}{l}{} &
  \multicolumn{1}{l}{Forgetting} &
  25.00 ± 0.3 &
  27.80 ± 1.1 &
  33.27 ± 2.0 &
  37.80 ± 1.0 &
  37.44 ± 2.2 &
  37.45 ± 1.7 \\
\multicolumn{1}{l}{} &
  \multicolumn{1}{l}{GraphCut} &
  24.04 ± 0.7 &
  30.45 ± 0.2 &
  33.31 ± 0.3 &
  35.76 ± 3.2 &
  38.03 ± 0.8 &
  37.45 ± 1.7 \\ \hline
\multicolumn{1}{l}{\multirow{5}{*}{ER~\citep{er}}} &
  \multicolumn{1}{l}{Random} &
  25.23 ± 0.3 &
  31.58 ± 3.0 &
  37.64 ± 1.4 &
  39.25 ± 1.3 &
  40.66 ± 2.0 &
  39.53 ± 1.6 \\
\multicolumn{1}{l}{} &
  \multicolumn{1}{l}{Herding} &
  19.13 ± 5.4 &
  24.90 ± 6.3 &
  34.92 ± 4.0 &
  40.18 ± 2.1 &
  \textbf{41.19 ± 1.2} &
  39.53 ± 1.6 \\
\multicolumn{1}{l}{} &
  \multicolumn{1}{l}{Uncertainty} &
  25.77 ± 4.6 &
  31.63 ± 4.3 &
  36.61 ± 1.5 &
  41.14 ± 0.4 &
  39.69 ± 1.4 &
  39.53 ± 1.6 \\
\multicolumn{1}{l}{} &
  \multicolumn{1}{l}{Forgetting} &
  29.53 ± 4.7 &
  33.97 ± 3.8 &
  36.96 ± 3.4 &
  40.58 ± 0.7 &
  39.92 ± 2.5 &
  39.53 ± 1.6 \\
\multicolumn{1}{l}{} &
  \multicolumn{1}{l}{GraphCut} &
  32.99 ± 8.7 &
  38.22 ± 6.4 &
  39.55 ± 3.5 &
  39.61 ± 2.6 &
  39.97 ± 0.6 &
  39.53 ± 1.6 \\ \hline
\multicolumn{1}{l}{\multirow{5}{*}{LwF~\citep{lwf}}} &
  \multicolumn{1}{l}{Random} &
  11.39 ± 1.0 &
  15.38 ± 1.3 &
  20.26 ± 1.3 &
  22.93 ± 2.1 &
  \textbf{23.91 ± 1.2} &
  22.82 ± 1.4 \\
\multicolumn{1}{l}{} &
  \multicolumn{1}{l}{Herding} &
  \hspace{5pt}3.67 ± 1.3 &
  \hspace{5pt}6.22 ± 0.1 &
  12.43 ± 2.0 &
  17.09 ± 4.6 &
  18.08 ± 4.5 &
  22.82 ± 1.4 \\
\multicolumn{1}{l}{} &
  \multicolumn{1}{l}{Uncertainty} &
  \hspace{5pt}9.55 ± 0.5 &
  12.17 ± 1.8 &
  15.54 ± 2.8 &
  18.72 ± 5.0 &
  18.00 ± 4.2 &
  22.82 ± 1.4 \\
\multicolumn{1}{l}{} &
  \multicolumn{1}{l}{Forgetting} &
  \hspace{5pt}9.93 ± 1.3 &
  12.75 ± 2.7 &
  15.18 ± 2.9 &
  17.99 ± 4.5 &
  18.28 ± 4.4 &
  22.82 ± 1.4 \\
\multicolumn{1}{l}{} &
  \multicolumn{1}{l}{GraphCut} &
  \hspace{5pt}8.17 ± 0.3 &
  10.37 ± 1.4 &
  15.56 ± 3.4 &
  17.26 ± 4.1 &
  18.00 ± 4.9 &
  22.82 ± 1.4 \\ \hline
\end{tabular}%
}
\end{table}

\paragraph{FOSTER benefits from more samples.}
FOSTER's primary objective is to identify critical elements that were potentially overlooked or misinterpreted by the original model during the learning process. For instance, in the initial stages of learning, certain features may have been deemed less significant than others. However, as the model progresses and encounters new concepts, previously redundant features may become crucial. FOSTER addresses these dynamics by employing a feature-boosting mechanism, which aims to highlight the evolving importance of features over time.
However, this mechanism may necessitate access to more samples to effectively capture the intricate relationships between features. Consequently, training with the full dataset enables the model to develop a more comprehensive understanding of the underlying patterns and correlations among the features. 

\begin{figure*}[h]
  \centering
  \includegraphics[width=0.8\textwidth]{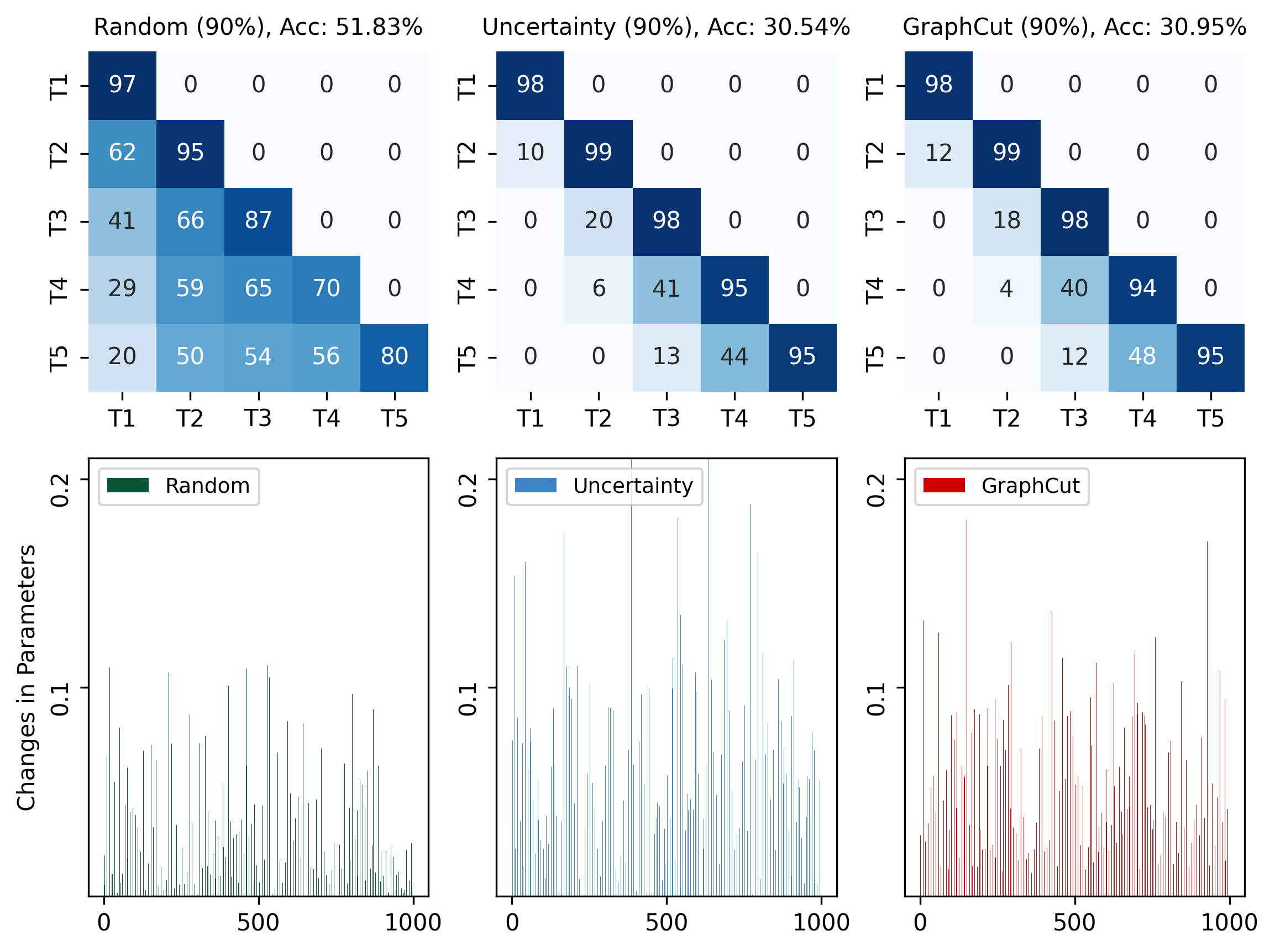}
    \caption{Accuracy [\%] after each learning step on LwF \textit{(above)}, reveals that Random selection demonstrates relatively less forgetting while effectively learning. This is due to the abrupt parameter changes. For example, on the last layer between consecutive tasks \textit{(below)}, Uncertainty and GraphCut abruptly shift the parameters.}
  \label{fig:weight_change}
\end{figure*}

\paragraph{LwF exhibits abrupt weight changes when trained with a coreset.}
Sophisticated coreset selection approaches do not yield performance advantages in LwF. Surprisingly, learning from a random samples appears to drive improvements instead. To understand this phenomenon, we conduct an in-depth investigation, focusing on the performance after each task, as illustrated in Figure \ref{fig:weight_change}.
Our analysis shows that LwF trained with more advanced coreset selection methods, such as Uncertainty and GraphCut, demonstrate superior adaptability to the current task. However, this enhanced adaptability comes at a cost of catastrophic forgetting. To unravel the root cause of this forgetting phenomenon, we examine the changes in model parameters between consecutive tasks.
We found that Uncertainty and GraphCut induce abrupt changes in the parameters, whereas it is comparatively smaller with randomly selected samples. 
This suggests that the traditional regularization methods may not be as effective as replay-based approaches when considering coreset utilization.

\begin{table}[h]
\caption{Accuracy [\%] of CIL models across various coreset fractions and selections on \textbf{Split-ImageNet100}. Learning from coreset samples enhances the performance, except FOSTER and LwF. The best results are highlighted in bold if coreset outperforms training with all samples.}
\label{tab:imagenet_subset}
\resizebox{\textwidth}{!}{%
\begin{tabular}{llllllll}
\hline
 &
  Fraction &
  \multicolumn{1}{c}{10\%} &
  \multicolumn{1}{c}{20\%} &
  \multicolumn{1}{c}{50\%} &
  \multicolumn{1}{c}{80\%} &
  \multicolumn{1}{c}{90\%} &
  \multicolumn{1}{c}{100\%} \\ \hline
\multicolumn{1}{l}{\multirow{5}{*}{DER~\citep{der}}} &
  \multicolumn{1}{l}{Random} &
  19.89 ± 2.3 &
  32.70 ± 1.5 &
  42.45 ± 0.6 &
  52.61 ± 1.8 &
  53.12 ± 1.0 &
  55.03 ± 1.2 \\
\multicolumn{1}{l}{} &
  \multicolumn{1}{l}{Herding} &
  18.30 ± 1.2 &
  29.83 ± 0.6 &
  44.77 ± 0.8 &
  53.59 ± 0.3 &
  55.52 ± 0.1 &
  55.03 ± 1.2 \\
\multicolumn{1}{l}{} &
  \multicolumn{1}{l}{Uncertainty} &
  27.08 ± 0.5 &
  36.92 ± 0.9 &
  49.84 ± 0.4 &
  55.10 ± 0.2 &
  \textbf{56.46 ± 0.6} &
  55.03 ± 1.2 \\
\multicolumn{1}{c}{} &
  \multicolumn{1}{l}{Forgetting} &
  32.69 ± 2.1 &
  40.21 ± 1.3 &
  50.27 ± 0.9 &
  55.15 ± 0.7 &
  55.60 ± 0.8 &
  55.03 ± 1.2 \\
\multicolumn{1}{c}{} &
  \multicolumn{1}{l}{GraphCut} &
  32.91 ± 0.7 &
  38.90 ± 0.4 &
  50.12 ± 0.8 &
  54.71 ± 0.3 &
  55.81 ± 0.1 &
  55.03 ± 1.2 \\ \hline
\multicolumn{1}{l}{\multirow{5}{*}{FOSTER~\citep{foster}}} &
  \multicolumn{1}{l}{Random} &
  17.59 ± 1.3 &
  22.68 ± 0.8 &
  34.20 ± 3.8 &
  46.90 ± 4.1 &
  48.64 ± 4.2 &
  52.06 ± 0.4 \\
\multicolumn{1}{l}{} &
  \multicolumn{1}{l}{Herding} &
 \hspace{3pt}8.67 ± 0.1 &
  13.42 ± 0.2 &
  30.63 ± 1.7 &
  45.85 ± 1.0 &
  48.89 ± 0.1 &
  52.06 ± 0.4 \\
\multicolumn{1}{l}{} &
  \multicolumn{1}{l}{Uncertainty} &
  \hspace{3pt}8.14 ± 0.1 &
  15.91 ± 0.5 &
  35.40 ± 0.5 &
  46.39 ± 0.5 &
  48.37 ± 0.5 &
  52.06 ± 0.4 \\
\multicolumn{1}{l}{} &
  \multicolumn{1}{l}{Forgetting} &
  11.62 ± 0.5 &
  18.71 ± 0.4 &
  35.26 ± 0.3 &
  46.95 ± 0.9 &
  49.45 ± 0.4 &
  52.06 ± 0.4 \\
\multicolumn{1}{l}{} &
  \multicolumn{1}{l}{GraphCut} &
  16.74 ± 0.5 &
  22.99 ± 0.1 &
  37.42 ± 0.4 &
  47.22 ± 0.4 &
  49.95 ± 0.9 &
  52.06 ± 0.4 \\ \hline
\multicolumn{1}{l}{\multirow{5}{*}{MEMO~\citep{memo}}} &
  \multicolumn{1}{l}{Random} &
  18.79 ± 0.1 &
  27.29 ± 0.2 &
  40.02 ± 1.7 &
  44.48 ± 0.2 &
  47.80 ± 1.9 &
  46.36 ± 1.0 \\
\multicolumn{1}{l}{} &
  \multicolumn{1}{l}{Herding} &
  18.15 ± 1.1 &
  26.08 ± 0.4 &
  37.71 ± 3.1 &
  46.76 ± 2.3 &
  47.94 ± 1.1 &
  46.36 ± 1.0 \\
\multicolumn{1}{l}{} &
  \multicolumn{1}{l}{Uncertainty} &
  20.22 ± 0.8 &
  26.94 ± 2.2 &
  39.39 ± 1.1 &
  45.90 ± 0.4 &
  \textbf{48.54 ± 0.2} &
  46.36 ± 1.0 \\
\multicolumn{1}{l}{} &
  \multicolumn{1}{l}{Forgetting} &
  24.40 ± 1.5 &
  33.16 ± 1.0 &
  41.86 ± 0.5 &
  45.57 ± 0.5 &
  47.19 ± 0.9 &
  46.36 ± 1.0 \\
\multicolumn{1}{l}{} &
  \multicolumn{1}{l}{GraphCut} &
  29.76 ± 1.8 &
  35.73 ± 1.1 &
  42.80 ± 1.9 &
  45.98 ± 2.8 &
  48.50 ± 1.3 &
  46.36 ± 1.0 \\ \hline
\multicolumn{1}{l}{\multirow{5}{*}{iCaRL~\citep{icarl}}} &
  \multicolumn{1}{l}{Random} &
  21.93 ± 0.7 &
  27.29 ± 0.5 &
  30.21 ± 3.7 &
  29.12 ± 1.9 &
  30.30 ± 1.6 &
  33.05 ± 1.8 \\
\multicolumn{1}{l}{} &
  \multicolumn{1}{l}{Herding} &
  20.80 ± 1.8 &
  24.29 ± 2.3 &
  30.92 ± 0.2 &
  33.23 ± 0.9 &
  34.04 ± 0.2 &
  33.05 ± 1.8 \\
\multicolumn{1}{l}{} &
  \multicolumn{1}{l}{Uncertainty} &
  22.52 ± 0.3 &
  22.37 ± 0.9 &
  32.67 ± 1.6 &
  33.03 ± 0.1 &
  34.76 ± 0.9 &
  33.05 ± 1.8 \\
\multicolumn{1}{l}{} &
  \multicolumn{1}{l}{Forgetting} &
  26.38 ± 0.1 &
  28.35 ± 0.8 &
  31.85 ± 0.7 &
  33.80 ± 0.6 &
  34.77 ± 2.7 &
  33.05 ± 1.8 \\
\multicolumn{1}{l}{} &
  \multicolumn{1}{l}{GraphCut} &
  33.04 ± 0.6 &
  35.10 ± 0.6 &
  34.87 ± 1.1 &
  \textbf{35.19 ± 0.2} &
  31.29 ± 0.3 &
  33.05 ± 1.8 \\ \hline
\multicolumn{1}{l}{\multirow{5}{*}{ER~\citep{er}}} &
  \multicolumn{1}{l}{Random} &
  20.19 ± 0.1 &
  25.84 ± 2.7 &
  30.47 ± 2.0 &
  29.14 ± 1.0 &
  30.81 ± 0.6 &
  34.23 ± 4.2 \\
\multicolumn{1}{l}{} &
  \multicolumn{1}{l}{Herding} &
  20.21 ± 0.1 &
  24.56 ± 0.8 &
  29.81 ± 1.1 &
  31.92 ± 0.4 &
  33.68 ± 0.7 &
  34.23 ± 4.2 \\
\multicolumn{1}{l}{} &
  \multicolumn{1}{l}{Uncertainty} &
  20.82 ± 0.6 &
  23.08 ± 0.6 &
  29.23 ± 0.5 &
  29.35 ± 1.1 &
  30.74 ± 1.4 &
  34.23 ± 4.2 \\
\multicolumn{1}{l}{} &
  \multicolumn{1}{l}{Forgetting} &
  24.85 ± 0.6 &
  28.32 ± 1.4 &
  29.03 ± 0.2 &
  32.85 ± 0.4 &
  31.74 ± 2.1 &
  34.23 ± 4.2 \\
\multicolumn{1}{l}{} &
  \multicolumn{1}{l}{GraphCut} &
  30.13 ± 1.0 &
  30.52 ± 0.2 &
  \textbf{34.83 ± 0.6} &
  32.05 ± 1.5 &
  32.16 ± 0.5 &
  34.23 ± 4.2 \\ \hline
\multicolumn{1}{l}{\multirow{5}{*}{LwF~\citep{lwf}}} &
  \multicolumn{1}{l}{Random} &
  \hspace{3pt}9.25 ± 0.1 &
  11.22 ± 0.7 &
  15.88 ± 0.8 &
  16.27 ± 1.1 &
  \textbf{16.52 ± 0.5} &
  16.46 ± 1.8 \\
\multicolumn{1}{l}{} &
  \multicolumn{1}{l}{Herding} &
  \hspace{3pt}5.70 ± 0.5 &
  \hspace{3pt}7.65 ± 1.1 &
  10.70 ± 0.1 &
  11.33 ± 0.2 &
  11.64 ± 0.2 &
  16.46 ± 1.8 \\
\multicolumn{1}{l}{} &
  \multicolumn{1}{l}{Uncertainty} &
  \hspace{3pt}7.84 ± 0.1 &
  \hspace{3pt}8.07 ± 0.1 &
  11.27 ± 0.2 &
  11.41 ± 0.1 &
  11.51 ± 0.3 &
  16.46 ± 1.8 \\
\multicolumn{1}{l}{} &
  \multicolumn{1}{l}{Forgetting} &
  \hspace{3pt}7.38 ± 0.2 &
  10.01 ± 0.1 &
  11.60 ± 0.2 &
  12.15 ± 0.1 &
  12.57 ± 0.3 &
  16.46 ± 1.8 \\
\multicolumn{1}{l}{} &
  \multicolumn{1}{l}{GraphCut} &
  \hspace{3pt}7.41 ± 0.2 &
  \hspace{3pt}9.29 ± 0.8 &
  10.77 ± 0.5 &
  12.06 ± 0.2 &
  12.88 ± 0.1 &
  16.46 ± 1.8 \\ \hline
\end{tabular}}
\end{table}

\subsection{Incremental performance increases because models forget less}
\label{section 5.2}
The performance improvements observed in class-incremental learners when trained on coreset samples can be attributed to several factors:

(i) First, coreset samples are carefully selected to represent the most informative subset of the data, thereby reducing redundancy and focusing on critical information. This strategy enhances the model's capacity for retention of essential information while minimizing the risk of overfitting to less relevant data points. In other words, this allows more focused exposure to relevant data and develops robust representations that consolidate the acquired knowledge better, leading to improved performance in class-incremental learning scenarios.

(ii) Second, sample selection before training is also crucial in enhancing the data quality utilized during the replay or memory construction phase in continual learning. By filtering out potentially irrelevant or redundant data points beforehand, it ensures that only the most informative and representative samples are stored in memory. This contributes to enhanced retention or consolidation of learned knowledge from previous tasks over time by focusing on key patterns and relationships.

\begin{figure*}[h]
  \centering
  \begin{subfigure}[b]{0.5\textwidth}
    \centering
    \includegraphics[width=\textwidth]{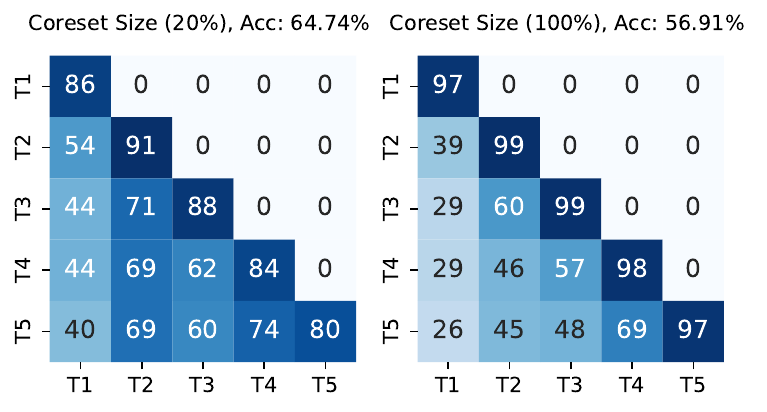}
    \caption{DER}
    \label{fig:der}
  \end{subfigure}%
  \begin{subfigure}[b]{0.5\textwidth}
    \centering
    \includegraphics[width=\textwidth]{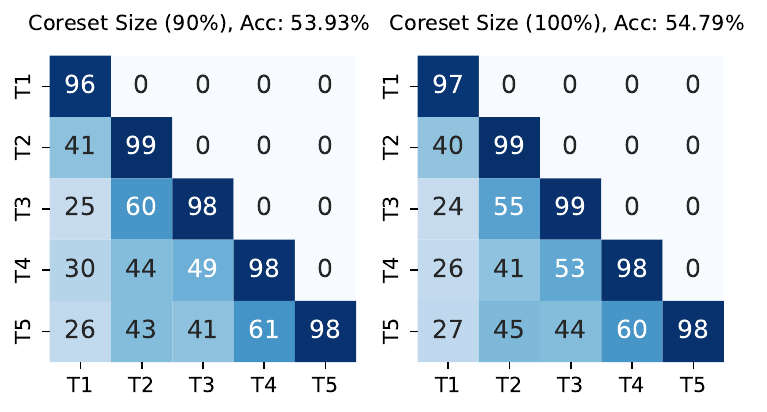}
    \caption{FOSTER}
    \label{fig:foster}
  \end{subfigure}%
  
  \begin{subfigure}[b]{0.5\textwidth}
    \centering
    \includegraphics[width=\textwidth]{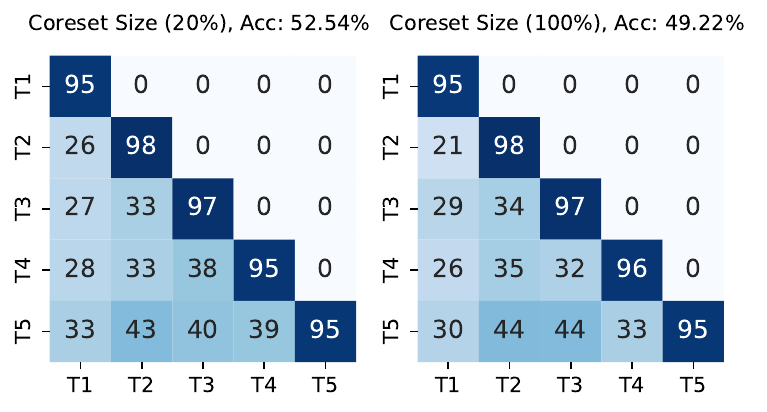}
    \caption{MEMO}
    \label{fig:memo}
  \end{subfigure}%
  \begin{subfigure}[b]{0.5\textwidth}
    \centering
    \includegraphics[width=\textwidth]{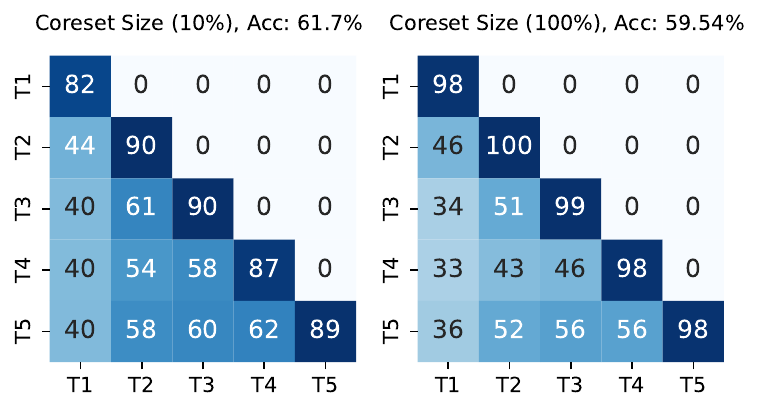}
    \caption{iCaRL}
    \label{fig:icarl}
  \end{subfigure}%
  
  \begin{subfigure}[b]{0.5\textwidth}
    \centering
    \includegraphics[width=\textwidth]{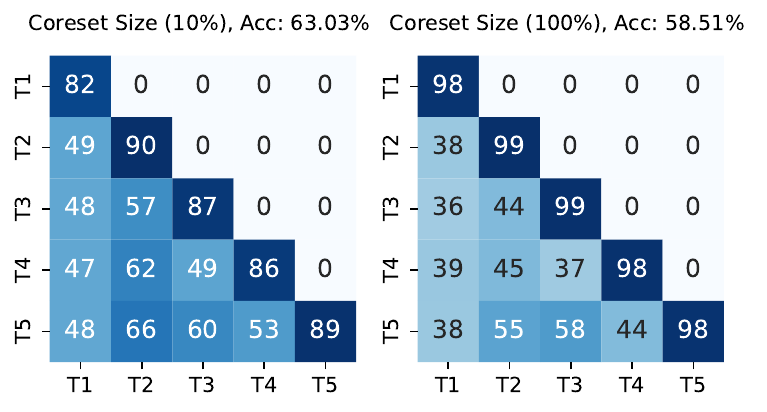}
    \caption{ER}
    \label{fig:er}
  \end{subfigure}%
  \begin{subfigure}[b]{0.5\textwidth}
    \centering
    \includegraphics[width=\textwidth]{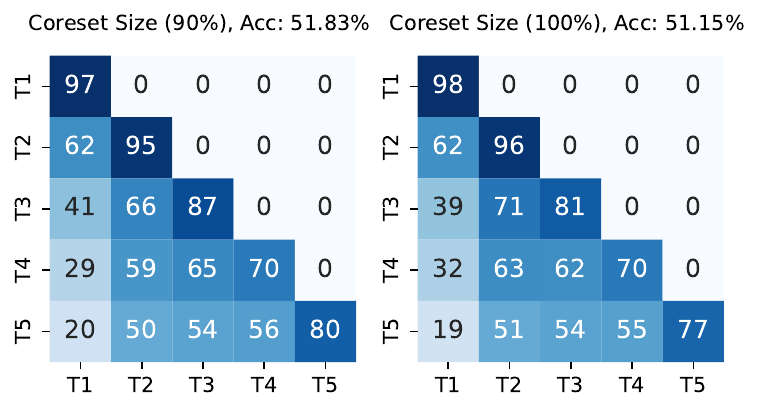}
    \caption{LwF}
    \label{fig:lwf}
  \end{subfigure}%
  \caption{Accuracy [\%] of each task after every learning session on different class-incremental learning methods with Split-CIFAR10. This comparison includes the performance using all samples \textit{vs.} the best performing coreset selection, which may involve different coreset fractions. The underlying reason for the improved accuracy is attributed to reduced forgetting.}
  \label{fig:confusion_matrix}
\end{figure*}

Consequently; DER, iCaRL, and ER demonstrate noticeable improvement in knowledge retention learning when trained on coreset samples (see Figure \ref{fig:confusion_matrix}). These methods leverage the enhanced representativeness and diversity of coreset samples, reinforcing old knowledge retention while learning new ones. MEMO and LwF also benefit from training on coreset samples, albeit to a lesser extent.
FOSTER still appears to rely more heavily on learning from the complete dataset, maintaining consistent performance across tasks. This reaffirms that its learning strategy may be better suited to leveraging the full dataset rather than coreset samples as we discuss above. 
In the Appendix \ref{appendix-cifar100}, we also provide more details and share the accuracy per task after each learning session on Split-CIFAR100. Overall, our analysis indicates that the enhanced incremental performance with coreset selection is primarily attributed to knowledge retention.

\subsection{Models forget less due to preserved representations}
\label{section 5.3}

\begin{figure*}[h]
  \centering
  \includegraphics[width=\textwidth]{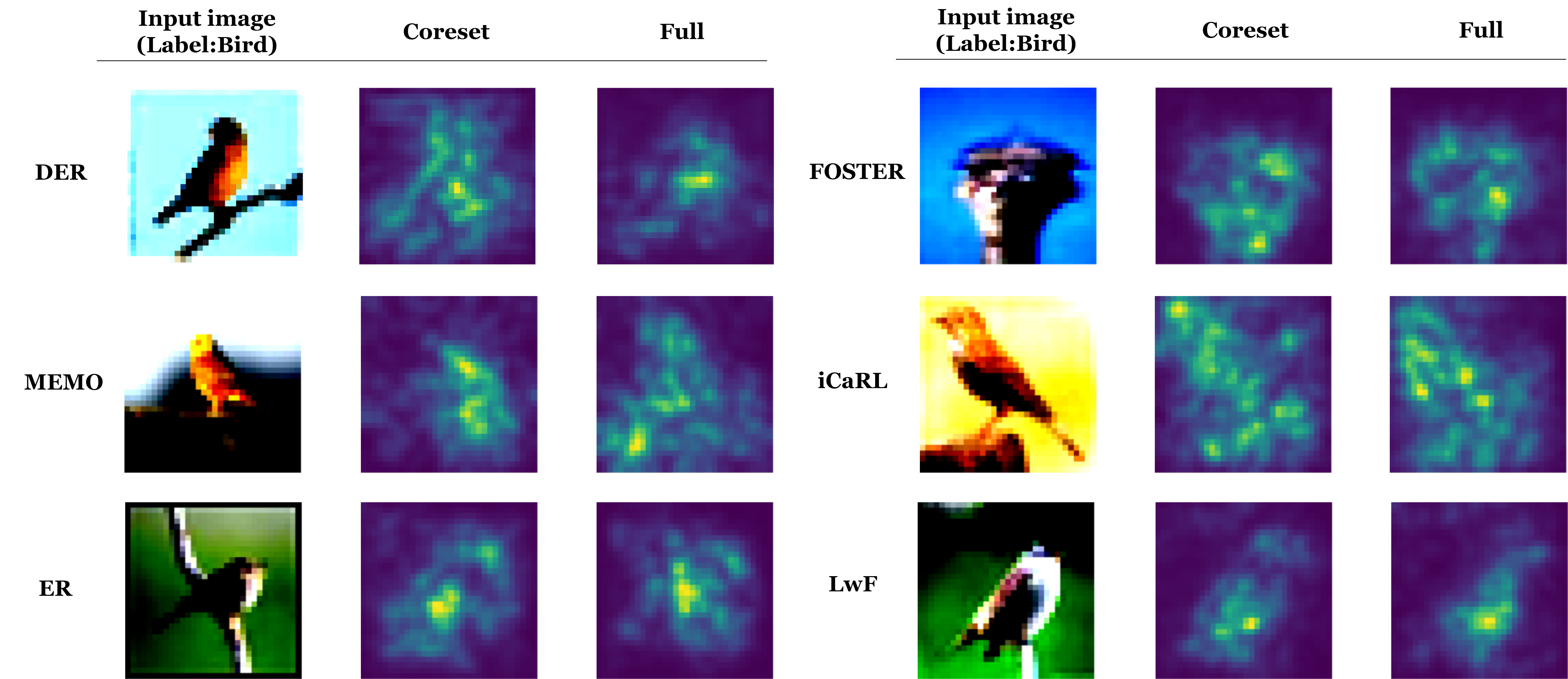}
    \caption{Saliency maps from the first encountered task after completing all learning sessions. Models trained with selected coresets exhibit enhanced perception capabilities in capturing the important parts of an input. Note that we select top performing coreset selection methods across different class-incremental learners.}
  \label{fig:saliency}
\end{figure*}

Here, we delve deeper into the key factor that drives enhanced knowledge retention. Specifically, we aim to explore how different class-incremental learners' perceptions evolved under different coreset methods and fractions. To achieve this, we generate saliency maps, as illustrated in Figure~\ref{fig:saliency}, with the objective of discerning where the model directed its attention after being trained with a coreset and compare against all data samples.
We find that models trained with the coresets exhibit a greater ability to retain focus on the object itself, effectively capturing the essence of the image. In contrast, models trained on all data samples tend to shift their focus to areas outside the main object. This insight sheds light on our earlier discussion regarding the model's knowledge retention or \textit{not} forgetting ability, and highlights that coreset selection gives more attention to relevant features.

\begin{figure*}[h]
  \centering
  \includegraphics[width=\textwidth]{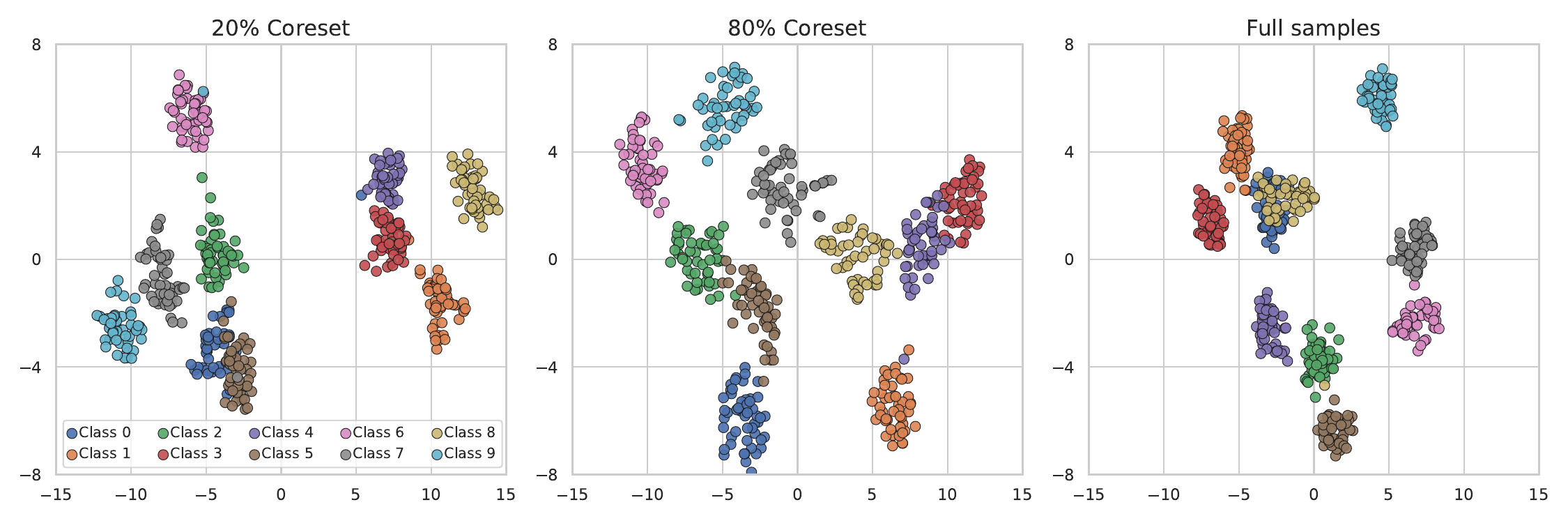}
    \caption{DER's representation of all classes on Split-CIFAR10 with varying coresets selected with GraphCut, compared to the full samples. When it is trained with coresets, it exhibits superior ability to distinct representations. 
    DER's representation of all classes with varying coresets selected with GraphCut, compared to the full samples. When it is trained with coresets, it exhibits superior ability to distinct representations.}
  \label{fig:tsne}
\end{figure*}

Furthermore, we investigate how the model's representation ability evolves as the coreset size changes, providing insights on the relationship between coreset composition and class separability. To illustrate this, in Figure~\ref{fig:tsne}, we employ DER to examine its representation of each class after completing all learning sessions. Notably, when using a smaller coreset, such as 20\%, the model demonstrates distinct separations between classes, effectively preserving boundaries between different categories. This suggests that with fewer, more concentrated samples, the model can maintain clearer distinctions.

However, as the coreset size increases, we observe a noticeable convergence in class representations, with boundaries between classes becoming less distinct. This trend suggests that larger coresets, while offering more data, may introduce redundancy or noise, causing overlap between classes and ultimately increasing the misclassification during inference. This phenomenon underscores the delicate balance between data quantity and quality, where more data does not necessarily translate into better generalization in class-incremental learning.

%% file: 6-conclusion.tex
\section{Conclusion}
\label{Conclusion}
Existing class-incremental learning approaches mostly use all available data during training yet all samples may not be equally informative and not need to go under the training process. In this unexplored area, we show that there is a potential in learning from important samples as the model's performance can be affected by the quality and quantity of the samples. Through extensive empirical analysis, we conclude three key findings that fundamentally reshape the understanding of CIL methodologies: (i) We find out that learning from coreset samples improves performance. (ii) We reveal that the observed performance improvement originated from improved knowledge retention over tasks by reducing redundancy and focusing on meaningful information. (iii) We elucidate how the model's perception evolves after completing all sessions and find that models trained with coresets show a greater ability to capture essential parts of the input and clear distinctions between classes. Overall, these findings, with underlying reasons, prove the significant impact of learning from coresets on continual learning. Our study contributes to a deeper understanding of current class-incremental learning methods and aims to inspire building more effective continual learners in real-world applications. Future analysis can include coreset utilization in online or blurry class-incremental learning setups.

%% file: 7-appendix.tex
\subsection{Results for Split-CIFAR100 with ResNet18}
\label{appendix-cifar100}
In Figure \ref{fig:cifar100_confusionmatrix}, we present the accuracy results of Split-CIFAR100 after each task for all class-incremental learning methods and compare the performance acquired using all samples of the dataset and the best-performing coreset results. These results may vary in terms of coreset size and selection method across different class-incremental learning methods.  Similar to observations made in Split-CIFAR10,  the underlying reason for the improved accuracy is attributed to reduced forgetting.

\renewcommand\thefigure{A}
 \begin{figure*}[h]
  \centering
   \begin{subfigure}[b]{0.5\textwidth}
     \centering
     \includegraphics[width=\textwidth]{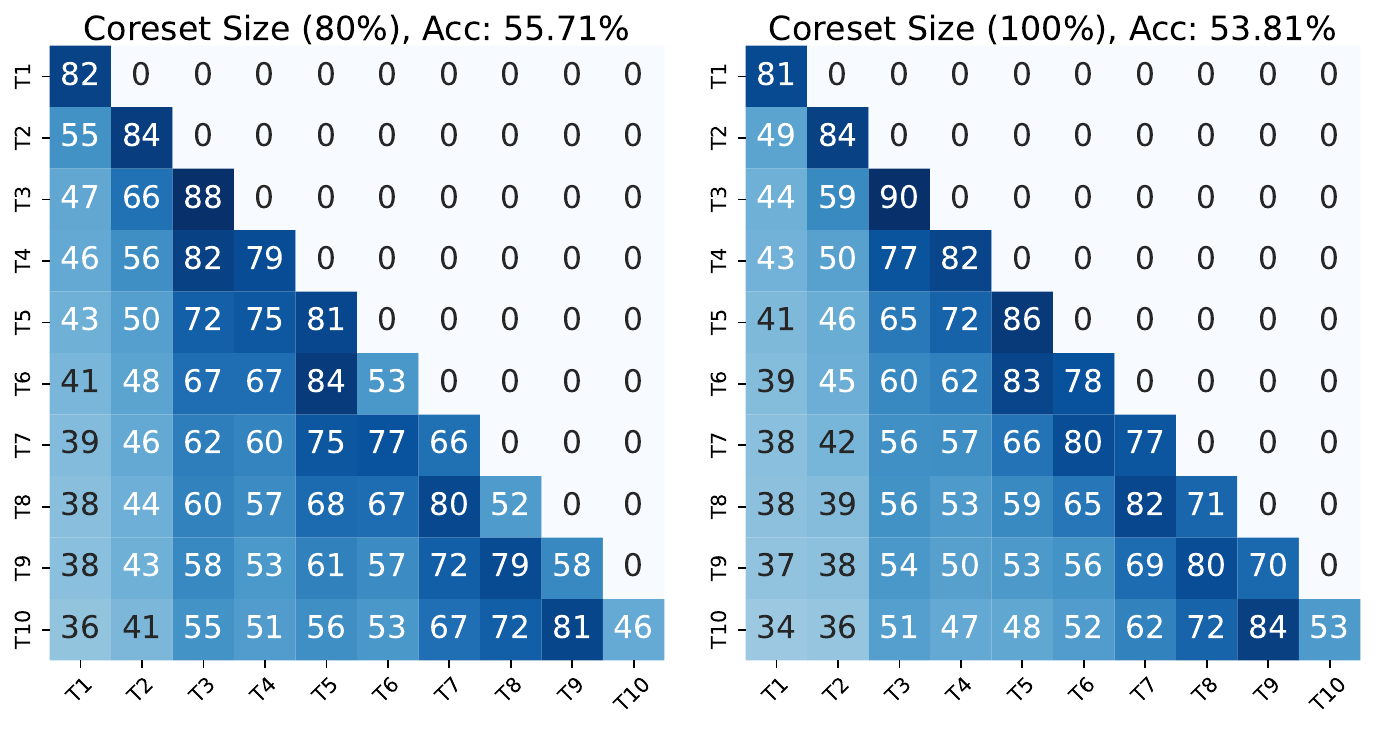}
     \caption{DER}
     \vspace{5pt}
     \label{fig:der_cifar100}
   \end{subfigure}%
   \begin{subfigure}[b]{0.5\textwidth}
     \centering
     \includegraphics[width=\textwidth]{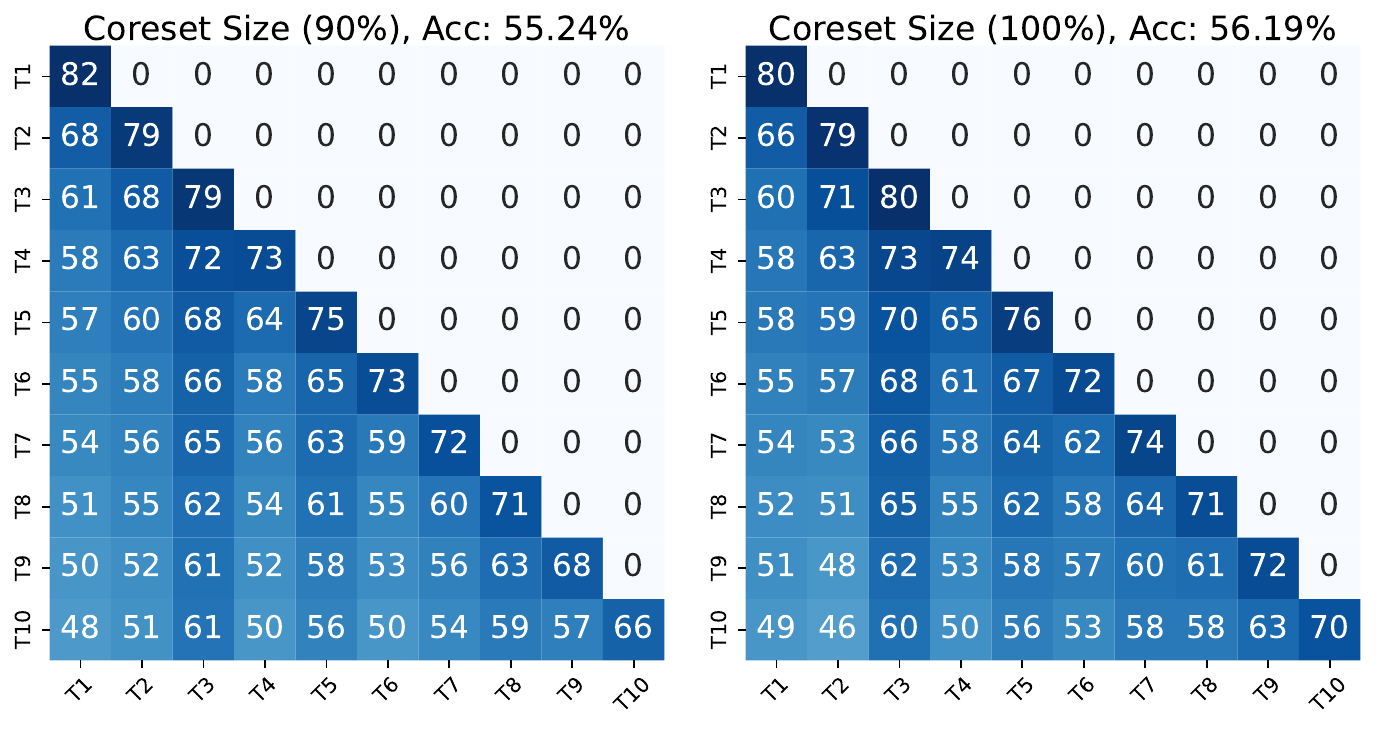}
     \caption{FOSTER}
     \vspace{5pt}
     \label{fig:foster_cifar100}
   \end{subfigure}\\
   \begin{subfigure}[b]{0.5\textwidth}
     \centering
     \includegraphics[width=\textwidth]{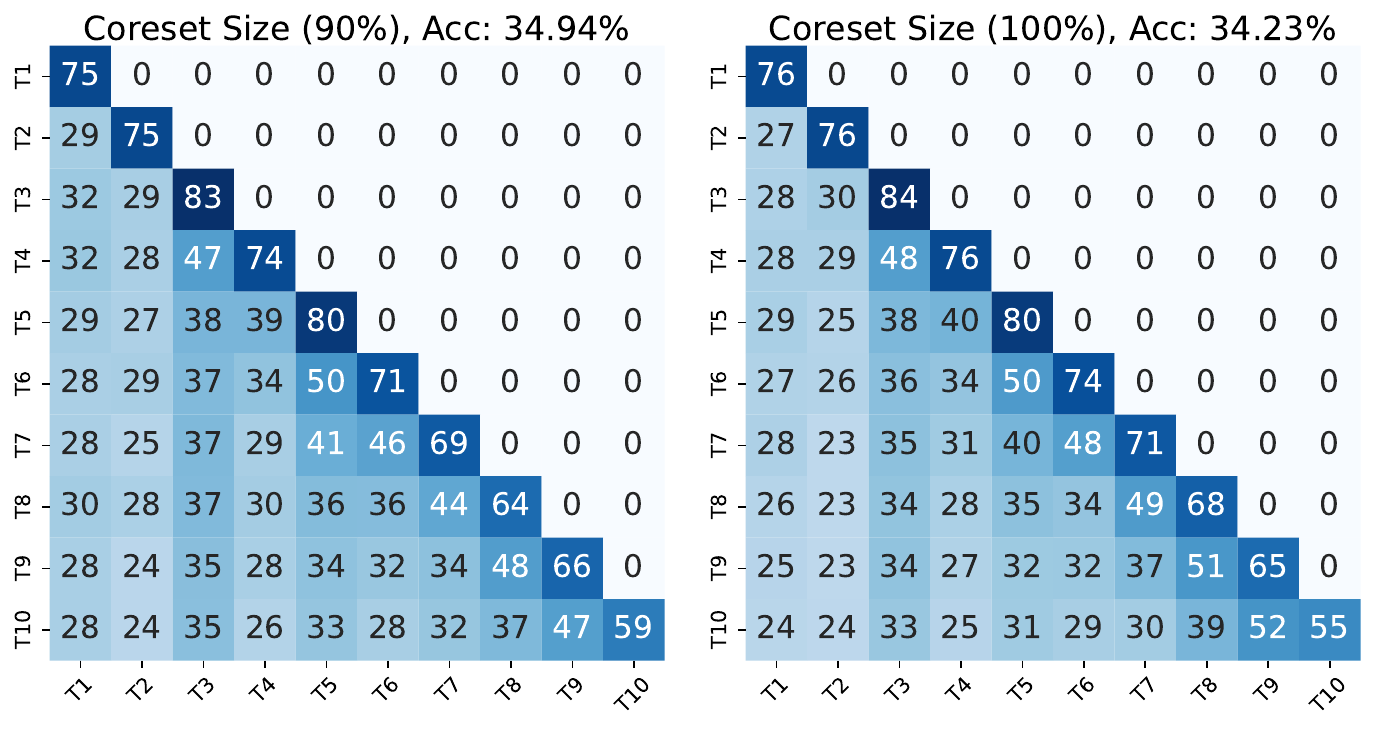}
     \caption{MEMO}
     \vspace{5pt}
     \label{fig:memo_cifar100}
   \end{subfigure}%
   \begin{subfigure}[b]{0.5\textwidth}
     \centering
     \includegraphics[width=\textwidth]{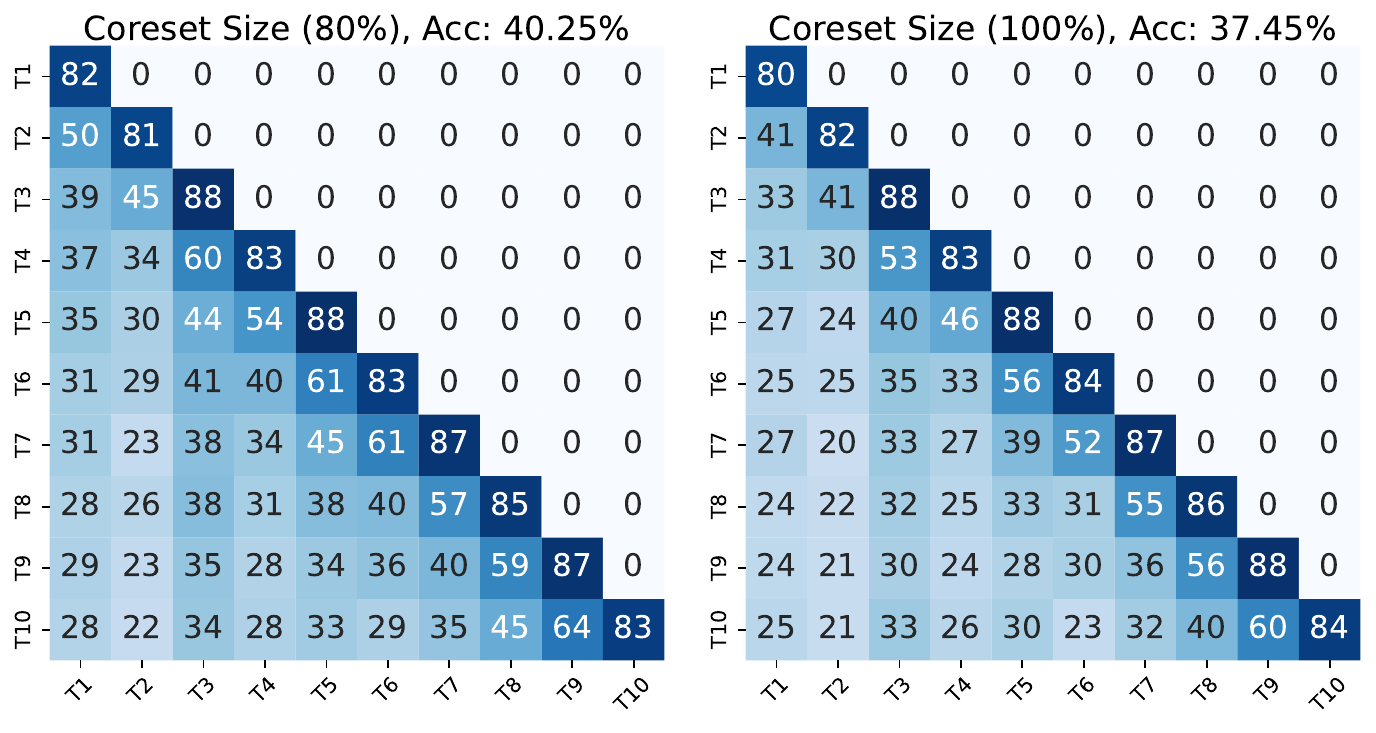}
     \caption{iCaRL}
     \vspace{5pt}
     \label{fig:icarl_cifar100}
   \end{subfigure}\\
  \begin{subfigure}[b]{0.5\textwidth}
     \centering
     \includegraphics[width=\textwidth]{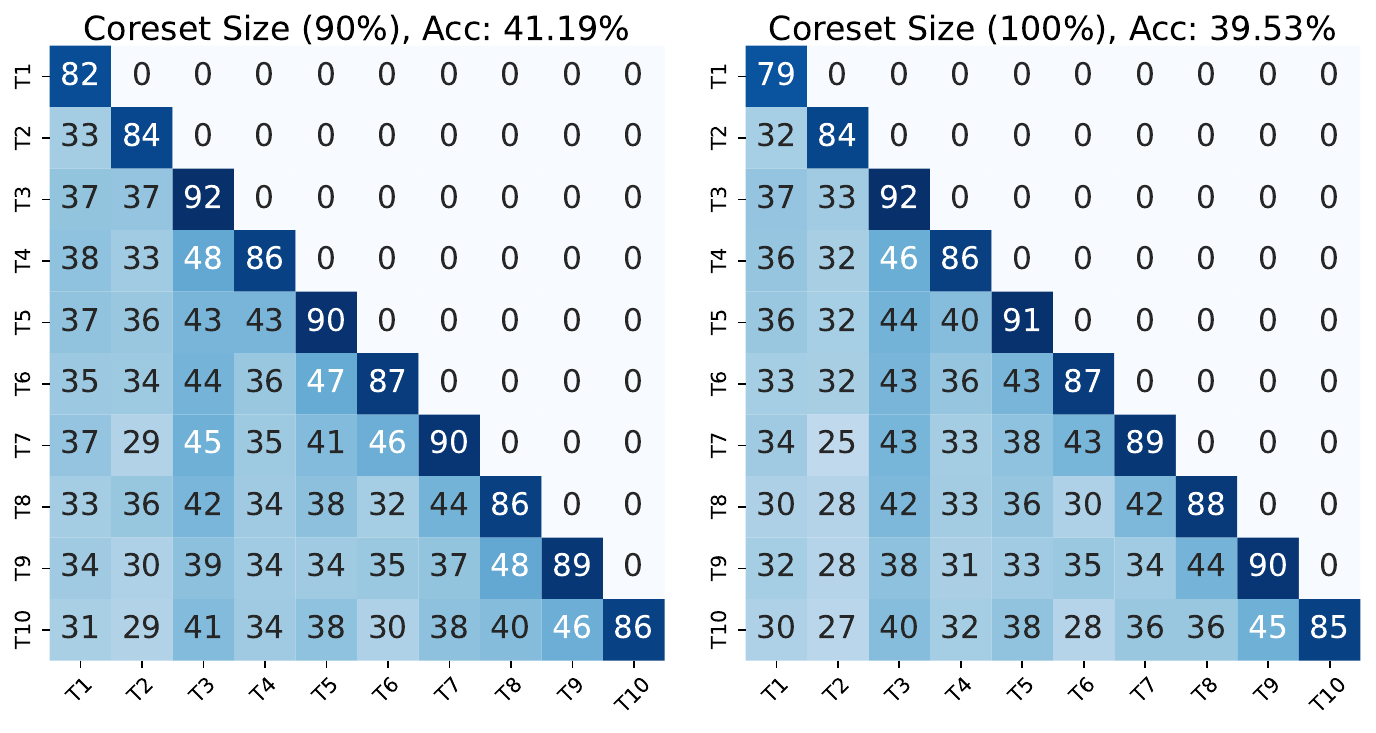}
     \caption{ER}
     \label{fig:er_cifar100}
   \end{subfigure}%
   \begin{subfigure}[b]{0.5\textwidth}
     \centering
     \includegraphics[width=\textwidth]{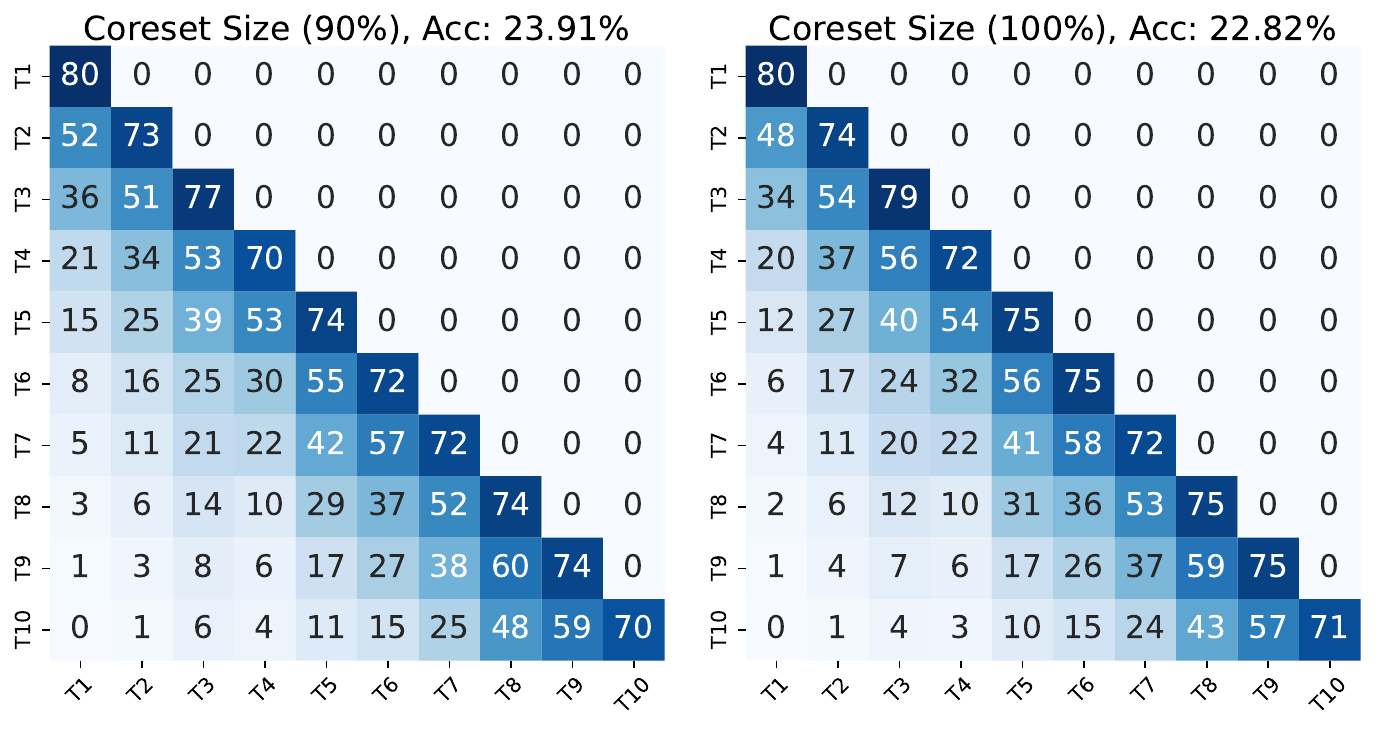}
    \caption{LwF}
     \label{fig:label_cifar100}
   \end{subfigure}%
   \caption{Accuracy [\%] of each task after every learning session on different class-incremental learning methods with Split-CIFAR100. Its results align with Split-CIFAR10 and again incremental performance improves due to better knowledge retention.}
   \label{fig:cifar100_confusionmatrix}
 \vskip -0.1in
 \end{figure*}

\subsection{Results for Pretrained Resnet18 and ViT}
\label{appendix-pretrained}
We investigate the impact of learning from important samples when prior information, such as a pretrained backbone, is available. It is worth noting that pretraining not always enhances performance, as pretrained parameters undergo continual fine-tuning with each task. Our experiments on Split-CIFAR10 and Split-CIFAR100 datasets consistently demonstrate that learning from coreset samples improves incremental performance when using ImageNet pretrained ResNet18 and ViT, aligning with our previous findings (Table \ref{tab:my-table-pretrained-cifar10}, Table \ref{tab:my-table-pretrained-cifar100}, and  Table \ref{tab:vit-cifar100}). While using ViT, we also apply a recent prompt-based method proposed for transformer architectures, called CODA-Prompt, to demonstrate that coreset selection is effective with such approaches.

\renewcommand\thetable{A}
\begin{table}[h]
\caption{Accuracy [\%] on \textbf{Split-CIFAR10} with an ImageNet \textbf{pretrained ResNet18} shows that training with coreset samples improves incremental performance.}
\label{tab:my-table-pretrained-cifar10}
\resizebox{\columnwidth}{!}{%
\begin{tabular}{llllllll}
\hline
 &
  Fraction &
  \multicolumn{1}{c}{10\%} &
  \multicolumn{1}{c}{20\%} &
  \multicolumn{1}{c}{50\%} &
  \multicolumn{1}{c}{80\%} &
  \multicolumn{1}{c}{90\%} &
  \multicolumn{1}{c}{100\%} \\ \hline
\multicolumn{1}{l}{\multirow{5}{*}{DER}} &
  \multicolumn{1}{l}{Random} &
  40.18 ± 5.28 &
  53.93 ± 3.36 &
  61.35 ± 2.37 &
  66.66 ± 2.36 &
  67.07 ± 2.51 &
  67.85 ± 3.30 \\
\multicolumn{1}{c}{} &
  \multicolumn{1}{l}{Herding} &
  57.35 ± 0.45 &
  61.48 ± 1.32 &
  65.84 ± 2.66 &
  68.68 ± 3.74 &
  \textbf{71.36 ± 1.48} &
  67.85 ± 3.30 \\
\multicolumn{1}{c}{} &
  \multicolumn{1}{l}{Uncertainty} &
  61.23 ± 0.14 &
  63.38 ± 0.40 &
  67.63 ± 1.37 &
  70.75 ± 2.71 &
  70.92 ± 2.08 &
  67.85 ± 3.30 \\
\multicolumn{1}{c}{} &
  \multicolumn{1}{l}{Forgetting} &
  61.00 ± 0.23 &
  65.02 ± 0.66 &
  67.86 ± 1.84 &
  71.72 ± 2.08 &
  69.67 ± 2.78 &
  67.85 ± 3.30 \\
\multicolumn{1}{c}{} &
  \multicolumn{1}{l}{GraphCut} &
  62.00 ± 2.03 &
  64.87 ± 1.92 &
  68.39 ± 0.98 &
  71.72 ± 1.65 &
  71.19 ± 2.77 &
  67.85 ± 3.30 \\ \hline
\multicolumn{1}{l}{\multirow{5}{*}{FOSTER}} &
  \multicolumn{1}{l}{Random} &
  42.82 ± 7.84 &
  46.24 ± 2.57 &
  60.15 ± 2.88 &
  57.89 ± 4.07 &
  58.32 ± 5.71 &
  57.85 ± 3.09 \\
\multicolumn{1}{l}{} &
  \multicolumn{1}{l}{Herding} &
  48.72 ± 4.27 &
  50.35 ± 2.35 &
  54.76 ± 4.46 &
  56.71 ± 2.53 &
  57.06 ± 3.39 &
  57.85 ± 3.09 \\
\multicolumn{1}{l}{} &
  \multicolumn{1}{l}{Uncertainty} &
  54.51 ± 1.48 &
  58.51 ± 2.97 &
  58.34 ± 3.54 &
  56.85 ± 4.81 &
  56.35 ± 3.38 &
  57.85 ± 3.09 \\
\multicolumn{1}{l}{} &
  \multicolumn{1}{l}{Forgetting} &
  52.26 ± 0.45 &
  55.52 ± 5.48 &
  57.61 ± 3.53 &
  57.65 ± 2.90 &
  55.98 ± 2.98 &
  57.85 ± 3.09 \\
\multicolumn{1}{l}{} &
  \multicolumn{1}{l}{GraphCut} &
  53.84 ± 3.70 &
  \textbf{59.27 ± 3.28} &
  58.04 ± 3.86 &
  57.57 ± 3.71 &
  56.09 ± 2.59 &
  57.85 ± 3.09 \\ \hline
\multicolumn{1}{l}{\multirow{5}{*}{MEMO}} &
  \multicolumn{1}{l}{Random} &
  37.49 ± 4.08 &
  43.77 ± 10.63 &
  48.74 ± 7.63 &
  53.90 ± 2.21 &
  59.34 ± 4.88 &
  55.65 ± 8.06 \\
\multicolumn{1}{l}{} &
  \multicolumn{1}{l}{Herding} &
  34.50 ± 7.48 &
  44.94 ± 12.11 &
  55.14 ± 7.53 &
  \textbf{62.84 ± 5.82} &
  61.34 ± 5.19 &
  55.65 ± 8.06 \\
\multicolumn{1}{l}{} &
  \multicolumn{1}{l}{Uncertainty} &
  43.02 ± 5.27 &
  50.06 ± 6.13 &
  54.55 ± 6.44 &
  61.21 ± 5.79 &
  62.00 ± 5.72 &
  55.65 ± 8.06 \\
\multicolumn{1}{l}{} &
  \multicolumn{1}{l}{Forgetting} &
  37.64 ± 4.28 &
  49.77 ± 8.80 &
  54.98 ± 6.70 &
  62.84 ± 5.78 &
  61.84 ± 6.93 &
  55.65 ± 8.06 \\
\multicolumn{1}{l}{} &
  \multicolumn{1}{l}{GraphCut} &
  47.23 ± 3.19 &
  52.04 ± 8.08 &
  55.96 ± 6.87 &
  61.57 ± 5.18 &
  61.37 ± 5.61 &
  55.65 ± 8.06 \\ \hline
\multicolumn{1}{l}{\multirow{5}{*}{iCaRL}} &
  \multicolumn{1}{l}{Random} &
  38.85 ± 0.13 &
  47.22 ± 7.77 &
  48.32 ± 3.87 &
  48.97 ± 3.02 &
  52.03 ± 5.62 &
  53.37 ± 5.94 \\
\multicolumn{1}{l}{} &
  \multicolumn{1}{l}{Herding} &
  53.52 ± 2.71 &
  55.21 ± 1.45 &
  53.68 ± 6.33 &
  55.42 ± 5.13 &
  55.38 ± 4.59 &
  53.37 ± 5.94 \\
\multicolumn{1}{l}{} &
  \multicolumn{1}{l}{Uncertainty} &
  53.72 ± 3.14 &
  56.03 ± 1.67 &
  52.81 ± 5.12 &
  56.82 ± 6.18 &
  54.73 ± 5.88 &
  53.37 ± 5.94 \\
\multicolumn{1}{l}{} &
  \multicolumn{1}{l}{Forgetting} &
  53.20 ± 0.90 &
  56.00 ± 4.88 &
  54.76 ± 5.06 &
  55.62 ± 5.33 &
  54.98 ± 6.39 &
  53.37 ± 5.94 \\
\multicolumn{1}{l}{} &
  \multicolumn{1}{l}{GraphCut} &
  \textbf{57.99 ± 2.41} &
  57.98 ± 3.45 &
  57.03 ± 3.85 &
  55.63 ± 4.50 &
  57.79 ± 5.47 &
  53.37 ± 5.94 \\ \hline
\multicolumn{1}{l}{\multirow{5}{*}{ER}} &
  \multicolumn{1}{l}{Random} &
  41.21 ± 2.43 &
  43.55 ± 6.68 &
  43.21 ± 5.02 &
  44.16 ± 6.60 &
  44.56 ± 6.71 &
  45.01 ± 5.56 \\
\multicolumn{1}{l}{} &
  \multicolumn{1}{l}{Herding} &
  38.28 ± 4.17 &
  41.91 ± 3.25 &
  47.91 ± 2.85 &
  44.76 ± 7.06 &
  43.17 ± 6.39 &
  45.01 ± 5.56 \\
\multicolumn{1}{l}{} &
  \multicolumn{1}{l}{Uncertainty} &
  36.23 ± 3.22 &
  40.28 ± 7.42 &
  42.19 ± 6.85 &
  44.01 ± 8.18 &
  43.81 ± 5.51 &
  45.01 ± 5.56 \\
\multicolumn{1}{l}{} &
  \multicolumn{1}{l}{Forgetting} &
  34.70 ± 3.03 &
  42.90 ± 5.67 &
  44.66 ± 6.07 &
  44.41 ± 6.35 &
  43.95 ± 6.01 &
  45.01 ± 5.56 \\
\multicolumn{1}{l}{} &
  \multicolumn{1}{l}{GraphCut} &
  \textbf{52.26 ± 3.93} &
  50.82 ± 4.91 &
  46.33 ± 5.23 &
  44.35 ± 7.20 &
  45.11 ± 7.77 &
  45.01 ± 5.56 \\ \hline
\multicolumn{1}{l}{\multirow{5}{*}{LwF}} &
  \multicolumn{1}{l}{Random} &
  30.80 ± 1.42 &
  41.67 ± 1.89 &
  45.95 ± 3.11 &
  51.04 ± 0.38 &
  \textbf{54.74 ± 0.44} &
  53.94 ± 0.79 \\
\multicolumn{1}{l}{} &
  \multicolumn{1}{l}{Herding} &
  17.65 ± 0.23 &
  21.74 ± 3.19 &
  26.41 ± 3.72 &
  29.85 ± 6.65 &
  31.53 ± 6.01 &
  53.94 ± 0.79 \\
\multicolumn{1}{l}{} &
  \multicolumn{1}{l}{Uncertainty} &
  25.21 ± 5.02 &
  26.38 ± 6.01 &
  27.76 ± 6.16 &
  30.68 ± 6.37 &
  32.13 ± 6.92 &
  53.94 ± 0.79 \\
\multicolumn{1}{l}{} &
  \multicolumn{1}{l}{Forgetting} &
  23.68 ± 1.81 &
  26.99 ± 5.19 &
  27.60 ± 5.33 &
  30.74 ± 5.98 &
  30.82 ± 6.81 &
  53.94 ± 0.79 \\
\multicolumn{1}{l}{} &
  \multicolumn{1}{l}{GraphCut} &
  26.45 ± 5.28 &
  25.23 ± 4.16 &
  27.79 ± 5.35 &
  31.05 ± 5.38 &
  31.78 ± 5.26 &
  53.94 ± 0.79 \\ \hline
\end{tabular}%
}
\end{table}

\renewcommand\thetable{B}
\begin{table}[h]
\caption{Accuracy [\%] on \textbf{Split-CIFAR100} with ImageNet \textbf{pretrained ResNet18}. Training with coreset samples improves the incremental performance also with a pretrained backbone.}
\label{tab:my-table-pretrained-cifar100}
\resizebox{\columnwidth}{!}{%
\begin{tabular}{llllllll}
\hline
 &
  Fraction &
  \multicolumn{1}{c}{10\%} &
  \multicolumn{1}{c}{20\%} &
  \multicolumn{1}{c}{50\%} &
  \multicolumn{1}{c}{80\%} &
  \multicolumn{1}{c}{90\%} &
  \multicolumn{1}{c}{100\%} \\ \hline
\multicolumn{1}{l}{\multirow{5}{*}{DER}} &
  \multicolumn{1}{l}{Random} &
  20.38 ± 3.27 &
  30.82 ± 0.76 &
  44.96 ± 0.28 &
  53.41 ± 1.96 &
  52.23 ± 0.84 &
  55.85  ± 0.38 \\
\multicolumn{1}{c}{} &
  \multicolumn{1}{l}{Herding} &
  16.33 ± 4.78 &
  22.13 ± 8.92 &
  47.52 ± 2.47 &
  55.51 ± 0.89 &
  56.74 ± 1.09 &
  55.85 ± 0.38 \\
\multicolumn{1}{c}{} &
  \multicolumn{1}{l}{Uncertainty} &
  30.03 ± 0.62 &
  40.53 ± 0.98 &
  52.21 ± 0.78 &
  56.94 ± 0.97 &
  \textbf{57.22 ± 0.59} &
  55.85 ± 0.38 \\
\multicolumn{1}{c}{} &
  \multicolumn{1}{l}{Forgetting} &
  30.08 ± 4.11 &
  37.48 ± 5.50 &
  51.88 ± 0.81 &
  56.18 ± 1.53 &
  56.16 ± 1.08 &
  55.85 ± 0.38 \\
\multicolumn{1}{c}{} &
  \multicolumn{1}{l}{GraphCut} &
  28.20 ± 1.64 &
  38.79 ± 1.66 &
  50.94 ± 1.59 &
  55.76 ± 0.68 &
  56.95 ± 1.77 &
  55.85 ± 0.38 \\ \hline
\multicolumn{1}{l}{\multirow{5}{*}{FOSTER}} &
  \multicolumn{1}{l}{Random} &
  16.25 ± 0.27 &
  19.71 ± 0.45 &
  34.21 ± 3.55 &
  50.80 ± 0.07 &
  50.65 ± 1.36 &
  56.63   ± 1.11 \\
\multicolumn{1}{l}{} &
  \multicolumn{1}{l}{Herding} &
  12.51 ± 0.03 &
  17.86 ± 1.39 &
  37.88 ± 1.58 &
  54.25 ± 2.37 &
  55.40 ± 2.13 &
  56.63 ± 1.11 \\
\multicolumn{1}{l}{} &
  \multicolumn{1}{l}{Uncertainty} &
  14.87 ± 1.03 &
  23.91 ± 0.86 &
  45.93 ± 1.50 &
  55.21 ± 2.26 &
  \textbf{56.65 ± 2.27} &
  56.63 ± 1.11 \\
\multicolumn{1}{l}{} &
  \multicolumn{1}{l}{Forgetting} &
  18.44 ± 0.72 &
  24.46 ± 1.84 &
  44.04 ± 0.33 &
  55.45 ± 2.08 &
  56.30 ± 1.21 &
  56.63 ± 1.11 \\
\multicolumn{1}{l}{} &
  \multicolumn{1}{l}{GraphCut} &
  17.87 ± 2.30 &
  22.10 ± 3.81 &
  44.94 ± 0.94 &
  55.51 ± 1.93 &
  56.60 ± 2.16 &
  56.63 ± 1.11 \\ \hline
\multicolumn{1}{l}{\multirow{5}{*}{MEMO}} &
  \multicolumn{1}{l}{Random} &
  17.21 ± 1.91 &
  25.29 ± 0.42 &
  38.54 ± 3.05 &
  43.16 ± 2.88 &
  46.32 ± 3.75 &
  46.70   ± 3.64 \\
\multicolumn{1}{l}{} &
  \multicolumn{1}{l}{Herding} &
  10.94 ± 0.72 &
  20.13 ± 0.21 &
  36.26 ± 0.94 &
  44.29 ± 0.75 &
  \textbf{46.87 ± 0.24} &
  46.70 ± 3.64 \\
\multicolumn{1}{l}{} &
  \multicolumn{1}{l}{Uncertainty} &
  17.85 ± 1.05 &
  24.54 ± 0.15 &
  37.92 ± 0.73 &
  44.87 ± 0.30 &
  46.10 ± 0.57 &
  46.70 ± 3.64 \\
\multicolumn{1}{l}{} &
  \multicolumn{1}{l}{Forgetting} &
  21.56 ± 0.52 &
  28.20 ± 0.51 &
  38.59 ± 1.06 &
  44.49 ± 0.88 &
  45.86 ± 0.58 &
  46.70 ± 3.64 \\
\multicolumn{1}{l}{} &
  \multicolumn{1}{l}{GraphCut} &
  27.60 ± 5.53 &
  33.44 ± 4.45 &
  40.38 ± 0.13 &
  44.54 ± 0.29 &
  45.60 ± 0.08 &
  46.70 ± 3.64 \\ \hline
\multicolumn{1}{l}{\multirow{5}{*}{iCaRL}} &
  \multicolumn{1}{l}{Random} &
  20.09 ± 0.72 &
  22.25 ± 0.93 &
  30.08 ± 0.04 &
  30.40 ± 1.16 &
  33.60 ± 0.66 &
  32.90   ± 0.80 \\
\multicolumn{1}{l}{} &
  \multicolumn{1}{l}{Herding} &
  18.46 ± 0.72 &
  24.80 ± 1.56 &
  32.74 ± 2.12 &
  34.70 ± 2.10 &
  34.74 ± 2.08 &
  32.90 ± 0.80 \\
\multicolumn{1}{l}{} &
  \multicolumn{1}{l}{Uncertainty} &
  22.70 ± 0.23 &
  27.82 ± 0.88 &
  32.68 ± 1.42 &
  33.44 ± 1.26 &
  34.04 ± 1.65 &
  32.90 ± 0.80 \\
\multicolumn{1}{l}{} &
  \multicolumn{1}{l}{Forgetting} &
  24.22 ± 0.69 &
  30.00 ± 1.38 &
  33.85 ± 2.05 &
  34.16 ± 2.72 &
  35.21 ± 2.10 &
  32.90 ± 0.80 \\
\multicolumn{1}{l}{} &
  \multicolumn{1}{l}{GraphCut} &
  28.88 ± 0.34 &
  30.93 ± 2.39 &
  \textbf{35.40 ± 1.56} &
  34.17 ± 0.96 &
  34.02 ± 1.47 &
  32.90 ± 0.80 \\ \hline
\multicolumn{1}{l}{\multirow{5}{*}{ER}} &
  \multicolumn{1}{l}{Random} &
  16.6 ± 3.59 &
  22.35 ± 0.04 &
  26.09 ± 0.34 &
  25.42 ± 0.10 &
  24.91 ± 0.16 &
  24.58 ± 0.46 \\
\multicolumn{1}{l}{} &
  \multicolumn{1}{l}{Herding} &
  15.2 ± 0.8 &
  19.9 ± 0.32 &
  25.16 ± 0.97 &
  25.94 ± 1.52 &
  25.30 ± 0.83 &
  24.58 ± 0.46 \\
\multicolumn{1}{l}{} &
  \multicolumn{1}{l}{Uncertainty} &
  14.4 ± 0.46 &
  17.56 ± 0.62 &
  22.78 ± 0.24 &
  24.04 ± 0.14 &
  25.58 ± 0.61 &
  24.58 ± 0.46 \\
\multicolumn{1}{l}{} &
  \multicolumn{1}{l}{Forgetting} &
  19.01 ± 0.63 &
  21.72 ± 0.14 &
  25.57 ± 0.69 &
  25.69 ± 0.89 &
  26.26 ± 1.55 &
  24.58 ± 0.46 \\
\multicolumn{1}{l}{} &
  \multicolumn{1}{l}{GraphCut} &
  27.01 ± 0.34 &
  \textbf{28.99 ± 1.63} &
  27.52 ± 0.57 &
  26.03 ± 1.43 &
  25.43 ± 0.86 &
  24.58 ± 0.46 \\ \hline
\multicolumn{1}{l}{\multirow{5}{*}{LwF}} &
  \multicolumn{1}{l}{Random} &
  10.39 ± 0.36 &
  12.63 ± 1.40 &
  20.69 ± 0.70 &
  22.78 ± 0.38 &
  \textbf{25.01 ± 0.46} &
  24.31 ± 0.57 \\
\multicolumn{1}{l}{} &
  \multicolumn{1}{l}{Herding} &
  4.15 ± 0.11 &
  5.44 ± 0.10 &
  9.47 ± 0.84 &
  13.11 ± 1.53 &
  13.77 ± 0.96 &
  24.31 ± 0.57 \\
\multicolumn{1}{l}{} &
  \multicolumn{1}{l}{Uncertainty} &
  7.42 ± 0.01 &
  9.15 ± 0.22 &
  11.00 ± 0.58 &
  13.29 ± 1.18 &
  14.46 ± 0.99 &
  24.31 ± 0.57 \\
\multicolumn{1}{l}{} &
  \multicolumn{1}{l}{Forgetting} &
  7.26 ± 0.24 &
  8.22 ± 0.17 &
  10.89 ± 0.86 &
  13.06 ± 1.14 &
  14.04 ± 0.94 &
  24.31 ± 0.57 \\
\multicolumn{1}{l}{} &
  \multicolumn{1}{l}{GraphCut} &
  6.59 ± 0.32 &
  7.23 ± 0.32 &
  11.13 ± 0.67 &
  13.21 ± 1.21 &
  13.65 ± 1.13 &
  24.31 ± 0.57 \\ \hline
\end{tabular}%
}
\end{table}

\renewcommand\thetable{C}
\begin{table}[]
\caption{Accuracy [\%] on \textbf{Split-CIFAR100} with ImageNet \textbf{pretrained ViT}. Training with coreset samples improves the incremental performance also with a pretrained backbone.}
\label{tab:vit-cifar100}
\resizebox{\columnwidth}{!}{%
\begin{tabular}{llllllll}
\hline
 &
  Fraction &
  \multicolumn{1}{c}{10\%} &
  \multicolumn{1}{c}{20\%} &
  \multicolumn{1}{c}{50\%} &
  \multicolumn{1}{c}{80\%} &
  \multicolumn{1}{c}{90\%} &
  \multicolumn{1}{c}{100\%} \\ \hline
\multicolumn{1}{l}{\multirow{5}{*}{DER}} &
  \multicolumn{1}{l}{Random} &
  61.51 ± 0.36 &
  61.88 ± 1.00 &
  64.39 ± 0.78 &
  63.12 ± 0.02 &
  64.10 ± 1.22 &
  60.83   ± 1.93 \\
\multicolumn{1}{c}{} &
  \multicolumn{1}{l}{Herding} &
  68.25 ± 1.44 &
  69.26 ± 1.15 &
  70.07 ± 0.15 &
  68.58 ± 1.03 &
  68.88 ± 1.92 &
  60.83 ± 1.93 \\
\multicolumn{1}{c}{} &
  \multicolumn{1}{l}{Uncertainty} &
  \textbf{74.44 ± 0.37} &
  71.30 ± 0.42 &
  69.68 ± 0.16 &
  68.92 ± 0.37 &
  70.28 ± 0.18 &
  60.83 ± 1.93 \\
\multicolumn{1}{c}{} &
  \multicolumn{1}{l}{Forgetting} &
  70.70 ± 2.70 &
  73.10 ± 0.55 &
  69.92 ± 1.23 &
  68.15 ± 0.63 &
  68.05 ± 0.09 &
  60.83 ± 1.93 \\
\multicolumn{1}{c}{} &
  \multicolumn{1}{l}{GraphCut} &
  72.58 ± 0.27 &
  72.29 ± 0.03 &
  69.70 ± 1.58 &
  68.88 ± 1.47 &
  68.37 ± 2.21 &
  60.83 ± 1.93 \\ \hline
\multicolumn{1}{l}{\multirow{5}{*}{FOSTER}} &
  \multicolumn{1}{l}{Random} &
  72.51 ± 2.67 &
  81.41 ± 0.67 &
  84.97 ± 0.56 &
  85.91 ± 0.28 &
  86.35 ± 0.42 &
  86.74 ± 0.30 \\
\multicolumn{1}{l}{} &
  \multicolumn{1}{l}{Herding} &
  68.84 ± 0.01 &
  78.87 ± 0.34 &
  83.68 ± 0.23 &
  85.41 ± 0.34 &
  85.58 ± 0.27 &
  86.74 ± 0.30 \\
\multicolumn{1}{l}{} &
  \multicolumn{1}{l}{Uncertainty} &
  77.10 ± 0.59 &
  82.68 ± 0.26 &
  85.17 ± 0.23 &
  86.03 ± 0.12 &
  85.83 ± 0.19 &
  86.74 ± 0.30 \\
\multicolumn{1}{l}{} &
  \multicolumn{1}{l}{Forgetting} &
  77.00 ± 1.53 &
  82.61 ± 0.30 &
  84.90 ± 0.39 &
  85.74 ± 0.33 &
  86.03 ± 0.24 &
  86.74 ± 0.30 \\
\multicolumn{1}{l}{} &
  \multicolumn{1}{l}{GraphCut} &
  74.64 ± 0.79 &
  79.72 ± 0.42 &
  84.14 ± 0.08 &
  85.09 ± 0.13 &
  85.68 ± 0.41 &
  86.74 ± 0.30 \\ \hline
\multicolumn{1}{l}{\multirow{5}{*}{MEMO}} &
  \multicolumn{1}{l}{Random} &
  14.84 ± 0.20 &
  17.87 ± 0.90 &
  23.74 ± 5.85 &
  27.24 ± 5.37 &
  30.07 ± 7.65 &
  36.12 ± 0.16 \\
\multicolumn{1}{l}{} &
  \multicolumn{1}{l}{Herding} &
  27.79 ± 1.15 &
  24.68 ± 1.79 &
  28.22 ± 2.03 &
  31.02 ± 0.66 &
  30.07 ± 0.45 &
  36.12 ± 0.16 \\
\multicolumn{1}{l}{} &
  \multicolumn{1}{l}{Uncertainty} &
  29.21 ± 1.47 &
  29.34 ± 1.07 &
  32.13 ± 0.76 &
  31.88 ± 2.99 &
  30.95 ± 0.10 &
  36.12 ± 0.16 \\
\multicolumn{1}{l}{} &
  \multicolumn{1}{l}{Forgetting} &
  35.14 ± 1.79 &
  31.72 ± 0.71 &
  29.29 ± 1.46 &
  31.47 ± 2.11 &
  31.00 ± 2.94 &
  36.12 ± 0.16 \\
\multicolumn{1}{l}{} &
  \multicolumn{1}{l}{GraphCut} &
  33.74 ± 1.66 &
  32.46 ± 2.07 &
  33.45 ± 3.05 &
  30.67 ± 3.23 &
  28.38 ± 2.63 &
  36.12 ± 0.16 \\ \hline
\multicolumn{1}{l}{\multirow{5}{*}{iCaRL}} &
  \multicolumn{1}{l}{Random} &
  71.24 ± 1.50 &
  71.79 ± 2.62 &
  70.62 ± 1.56 &
  68.30 ± 1.72 &
  68.79 ± 2.38 &
  66.03 ± 0.61 \\
\multicolumn{1}{l}{} &
  \multicolumn{1}{l}{Herding} &
  68.34 ± 0.21 &
  69.85 ± 0.25 &
  71.11 ± 0.48 &
  70.72 ± 0.41 &
  69.09 ± 0.59 &
  66.03 ± 0.61 \\
\multicolumn{1}{l}{} &
  \multicolumn{1}{l}{Uncertainty} &
  74.88 ± 0.41 &
  74.11 ± 0.14 &
  70.61 ± 0.13 &
  70.99 ± 0.34 &
  69.20 ± 0.45 &
  66.03 ± 0.61 \\
\multicolumn{1}{l}{} &
  \multicolumn{1}{l}{Forgetting} &
  73.21 ± 0.58 &
  73.51 ± 0.40 &
  71.91 ± 0.76 &
  70.74 ± 0.23 &
  70.61 ± 1.03 &
  66.03 ± 0.61 \\
\multicolumn{1}{l}{} &
  \multicolumn{1}{l}{GraphCut} &
  72.74 ± 4.08 &
  \textbf{73.68 ± 1.78} &
  71.72 ± 0.52 &
  73.05 ± 2.42 &
  73.59 ± 2.28 &
  66.03 ± 0.61 \\ \hline
\multicolumn{1}{l}{\multirow{5}{*}{ER}} &
  \multicolumn{1}{l}{Random} &
  69.52 ± 2.83 &
  73.54 ± 1.81 &
  73.59 ± 0.10 &
  73.36 ± 0.16 &
  72.39 ± 0.52 &
  67.95 ± 0.86 \\
\multicolumn{1}{l}{} &
  \multicolumn{1}{l}{Herding} &
  67.47 ± 1.53 &
  70.57 ± 0.20 &
  71.43 ± 1.43 &
  72.65 ± 0.60 &
  72.24 ± 0.16 &
  67.95 ± 0.86 \\
\multicolumn{1}{l}{} &
  \multicolumn{1}{l}{Uncertainty} &
  73.97 ± 0.25 &
  72.71 ± 1.94 &
  71.68 ± 0.38 &
  72.68 ± 0.84 &
  70.31 ± 0.37 &
  67.95 ± 0.86 \\
\multicolumn{1}{l}{} &
  \multicolumn{1}{l}{Forgetting} &
  71.32 ± 0.73 &
  71.31 ± 0.24 &
  71.50 ± 1.06 &
  72.00 ± 0.45 &
  72.09 ± 0.27 &
  67.95 ± 0.86 \\
\multicolumn{1}{l}{} &
  \multicolumn{1}{l}{GraphCut} &
  \textbf{76.59 ± 0.35} &
  76.39 ± 1.68 &
  74.87 ± 0.46 &
  70.09 ± 0.25 &
  70.69 ± 0.66 &
  67.95 ± 0.86 \\ \hline
\multicolumn{1}{l}{\multirow{5}{*}{LwF}} &
  \multicolumn{1}{l}{Random} &
  52.76 ± 2.27 &
  60.26 ± 2.62 &
  64.73 ± 1.56 &
  65.71 ± 0.70 &
  65.35 ± 0.85 &
  66.63 ± 1.41 \\
\multicolumn{1}{l}{} &
  \multicolumn{1}{l}{Herding} &
  22.99 ± 0.13 &
  24.44 ± 0.13 &
  27.57 ± 0.49 &
  29.46 ± 0.67 &
  31.10 ± 0.40 &
  66.63 ± 1.41 \\
\multicolumn{1}{l}{} &
  \multicolumn{1}{l}{Uncertainty} &
  25.17 ± 0.60 &
  26.27 ± 0.31 &
  28.78 ± 0.26 &
  30.19 ± 0.64 &
  30.31 ± 0.04 &
  66.63 ± 1.41 \\
\multicolumn{1}{l}{} &
  \multicolumn{1}{l}{Forgetting} &
  24.99 ± 0.29 &
  26.50 ± 0.18 &
  27.63 ± 0.74 &
  31.22 ± 0.77 &
  30.52 ± 0.44 &
  66.63 ± 1.41 \\
\multicolumn{1}{l}{} &
  \multicolumn{1}{l}{GraphCut} &
  23.32 ± 1.24 &
  25.84 ± 0.98 &
  29.53 ±1.47 &
  29.66 ± 0.45 &
  31.82 ± 0.75 &
  66.63 ± 1.41 \\ \hline
  \multicolumn{1}{l}{\multirow{5}{*}{CODA-Prompt}} &
  \multicolumn{1}{l}{Random} &
  78.99 ± 1.42 &
  81.62 ± 1.89 &
  84.01 ± 0.11 &
  84.64 ± 0.38 &
  85.45 ± 0.44 &
  85.37 ± 0.79 \\
\multicolumn{1}{l}{} &
  \multicolumn{1}{l}{Herding} &
  73.21 ± 1.23 &
  74.31 ± 1.19 &
  83.21 ± 0.72 &
  85.51 ± 0.65 &
  85.73 ± 0.01 &
  85.37 ± 0.79 \\
\multicolumn{1}{l}{} &
  \multicolumn{1}{l}{Uncertainty} &
  78.48 ± 1.02 &
  82.32 ± 1.01 &
  85.20 ± 0.16 &
  85.64 ± 0.37 &
  85.57 ± 0.92 &
  85.37 ± 0.79 \\
\multicolumn{1}{l}{} &
  \multicolumn{1}{l}{Forgetting} &
  78.30 ± 1.81 &
  82.48 ± 1.19 &
  84.73 ± 0.33 &
  85.73 ± 0.98 &
  86.33 ± 0.81 &
  85.37 ± 0.79 \\
\multicolumn{1}{l}{} &
  \multicolumn{1}{l}{GraphCut} &
  80.55 ± 1.28 &
  83.33 ± 1.16 &
  84.31 ± 0.35 &
  85.26 ± 0.38 &
  \textbf{86.34 ± 0.26} &
  85.37 ± 0.79 \\ \hline
\end{tabular}%
}
\end{table}